\documentclass{article}
\pdfoutput=1 
\usepackage{graphicx} 
\usepackage{arxiv}
\usepackage[numbers]{natbib}
\usepackage{placeins}
\usepackage{amsmath,amssymb}
\usepackage{longtable}
\usepackage{hyperref}
\renewcommand{\headeright}{}
\renewcommand{\undertitle}{}

\newcommand{\weburl}[1]{\href{https://#1}{\nolinkurl{#1}}}

\newcommand{\benchname}{OEIS Open}
\newcommand{\bench}{\textsc{\benchname}}
\newcommand{\benchlite}{\textsc{\benchname{} Lite}}

\title{\benchname{}: How many conjectures can language models turn into theorems?}
\author{Tom Adamczewski\\
Epoch AI\\
\texttt{tom@epoch.ai}}
\date{}

\begin{document}

\maketitle

\begin{abstract}
We construct \bench{}, a benchmark based on 492 open mathematical conjectures from the OEIS, formalized in Lean by Tsoukalas et al. Whereas these conjectures had previously been attempted only with a bespoke agent, our open-source evaluation code runs any generic language model (LM) against them, and is secure against LM cheating attempts. We find that LMs equipped with a minimal set of tools resolve 147 of these conjectures with a budget of \$50 per attempt, scoring 30\% on \bench{}.  \benchlite{} is a random subset of 100 conjectures for cheaper evaluation. When evaluated with a budget of \$200 per attempt, the best current LM scores 44\% on \benchlite{}. Giving LMs access to the mathematics literature via 476,000 papers from arXiv did not increase performance on \benchlite{}, and nor did using more sophisticated agent loops. The conjectures covered in this work are of uncertain mathematical significance, and most have likely received little previous attention.  
Nevertheless, our results  show that LMs can resolve open research conjectures autonomously and at modest cost.

\end{abstract}

\section{Introduction}

There have been several recent examples of AI systems solving open problems in mathematics.\footnote{On 20 May 2026, OpenAI announced an AI had disproved the unit distance conjecture \citep{openai2026unitdistance}; on July 19, Levent Alp\"{o}ge presented an explicit counterexample to the Jacobian conjecture in three-dimensional space, stating that it was discovered using Claude Fable 5 \citep{lee2026jacobian}; on 1 August OpenAI provided new results achieved by AI for ten problems in mathematics and theoretical computer science \citep{openai2026tenadvances}.} These are impressive demonstrations of AI's ability to contribute to research mathematics. However, they fall short of a systematic study of AI capabilities:

\begin{itemize}
    \item These reports do not disclose the universe of problems attempted by AI, so we do not know the success rate or average cost to solve such problems.
    \item The extent of guidance by human mathematicians is unclear. These projects are likely to involve mathematically sophisticated users, and the prompts and interactions behind the results have not been published, so human insight may have steered the search to an unknown degree.
    \item Different AI models are not systematically compared on the same problems, so we do not know when AI models first became capable of proving these results, and nor do we know whether some current models perform better than others.
\end{itemize}

The primary reason to benchmark AIs on open problems rather than difficult puzzles with known answers is to understand whether AI can push forward the frontier of a research field. Open problems also have a secondary advantage for the field of AI benchmarking: their solutions cannot leak into training corpora, at least until they are solved.

Existing benchmarks of open problems include \emph{HorizonMath} \citep[101 problems;][]{wang2026horizonmath} and \emph{FrontierMath: Open Problems} \citep[FM:OP, 50 problems;][]{epochai2026fmop}. Both make unsolved problems verifiable by restricting attention to problems with a generator--verifier gap. Candidate solutions (a concrete object, such as a graph, polynomial, or algorithm) are hard to find but cheap to check computationally. HorizonMath draws on three such classes: conjectured closed forms, optimization problems, and constructions of objects not known to exist. FM:OP sources problems from working mathematicians who must write a bespoke verifier program for each.

These strategies face challenges:

\begin{enumerate}
    \item They cannot be used to test most of research mathematics. The vast majority of open problems have no generator--verifier gap: they call for a proof of a general statement rather than the exhibition of a checkable object, and so cannot be expressed in this format at all.
    \item Passing the check is evidence rather than proof. HorizonMath describes accepted closed forms as ``best regarded as conjectures until proven'' and FM:OP explicitly allows verifiers that provide ``strong numerical evidence'' short of full proof.
    \item Soundness rests on hand-crafted verification code and filters.
    \begin{itemize}
        \item For example, HorizonMath accepts a conjectured closed form if it matches a high-precision numerical reference value. HorizonMath therefore uses an LLM judge to reject hard-coded constants and operations like numerical root-finding.
        \item FM:OP's bespoke verifiers are labor-intensive to formulate and implement. They are also error-prone: in July 2026, two problems were removed because their verifiers ``would not detect correct solutions with high enough fidelity''.
    \end{itemize}
    \item Computational checking is one-sided. A verifier can confirm that an exhibited object works, but if no such object exists, the benchmark task is unsolvable. Both benchmarks knowingly include problems that may have no solution of the required form. FM:OP estimates that 10--40\% of its problems are unsolvable.
\end{enumerate}

We instead require solutions to be formal proofs. Each conjecture is stated in the Lean proof assistant, and a model must prove either the conjecture or its negation. This addresses the first challenge above: eligibility requires only that a conjecture's statement be formalizable, not that its solutions be computationally checkable objects. The other challenges are avoided outright: an accepted proof is definitive rather than evidence, soundness rests on the Lean kernel rather than per-problem verification code, and false conjectures remain solvable tasks because the model may prove the negation.\footnote{One caveat: a conjecture, whether true or false, could in
principle be independent of Lean's axiomatic foundation, in which case neither it nor its negation is provable and the task is unsolvable.} Using formal proofs still has some downsides:

\begin{enumerate}
    \item Not all research mathematics can be covered: it must be possible to state conjectures with the definitions available in Mathlib (or short auxiliary definitions).
    \item Results reflect formalization ability as well as mathematical ability, since a model may find a correct argument yet fail to formalize it.
\end{enumerate}

There is a risk that formal conjectures are misformalized, meaning the Lean statement does not reflect the conjecture intended by the mathematician. But similar risks are present in the other approaches. A verifier may fail due to a range of pathologies, ranging from simple bugs to subtle and unforeseen false positives or false negatives. As described below, we use conjectures about integer sequences to reduce misformalization risk.

Current data does not let us reliably characterize the degree of mathematical interest these conjectures have received. While they are open conjectures proposed by mathematicians and approved by volunteer editors, it appears likely that most have received received little attention.

The benchmark and evaluation scaffold are released at \weburl{github.com/epoch-research/LeanOpenProblems}.

\section{Methods}

\subsection{Conjectures}
The conjectures were collected and formalized by \citet{tsoukalas2026apn}. They ``began with a corpus of 2649 open conjectures drawn from the OEIS'' (the Online Encyclopedia of Integer Sequences), ``prompted Gemini to select 500 problems that are non-trivial, mathematically interesting, not famous open problems, and good candidates for automated theorem-proving'', and ``used a Gemini-based agent to formalize them'' in Lean. Eight of the 500 were excluded for technical reasons.

Conjectures about integer sequences have relatively low misformalization risk because they generally involve only integers and elementary operations on them, rather than complicated mathematical objects whose formal statements rest on long chains of Mathlib definitions.

\subsection{Conjecture metadata}
\label{sec:metadata}

We also collected metadata on each conjecture: its provenance (who proposed it, and when) and the attention its sequence has received in the literature. The 492 conjectures concern 444 distinct OEIS sequences, some contributing several conjectures. For each sequence we fetched the full OEIS record and its complete revision history, then used GPT-5.5 to match each Lean conjecture to the OEIS text it formalizes and to identify its proposer and proposal date, distinguishing the proposer from users who merely edited or verified the entry. The proposal date almost always corresponds to the date the conjecture text first entered the database; the proposer is typically the sequence's author or a signed attribution in the entry's text. This yielded a proposer for 488 and a date for 489 of the 492 conjectures, with 443 extractions rated high-confidence by the model.

We measured literature attention to sequences in two ways: citations listed on the OEIS entry itself (its links and references, excluding OEIS-internal items such as b-files), and works in the OpenAlex database whose full text references the sequence.\footnote{We use the union of three OpenAlex full-text searches: (i) the bare A-number (e.g. ``A051293''), restricted to four mathematics subfields (Algebra and Number Theory; Discrete Mathematics and Combinatorics; Theoretical Computer Science; Computational Mathematics), since strings of that shape also occur as identifiers in other disciplines; (ii) the quoted URL ``oeis.org/A051293'', in any discipline; and (iii) the A-number co-occurring with the token ``OEIS'', in any discipline.} Both measures exclude references to Tsoukalas et al.'s own results \citep{tsoukalas2026apn}, which have already begun to appear on OEIS entries. Attention to a sequence is only a proxy for attention to a conjecture about that sequence: works that cite a sequence need not engage with the specific conjecture.

\subsection{Proof verification}
Following \citet{tsoukalas2026apn}, we accept a submission only if it passes SafeVerify, an open-source checker that ``checks the proof against the theorem specification and guards against environment exploits (e.g., axiom injection)''. SafeVerify is adapted from lean4checker, the external checker maintained by the Lean developers. Both replay compiled Lean through the kernel from scratch, but lean4checker certifies only that an environment is sound, whereas SafeVerify additionally certifies that the submission proves the target statement: each target declaration must be present with the same name, kind, and kernel type, and may use no axioms beyond the standard three (\texttt{propext}, \texttt{Quot.sound}, \texttt{Classical.choice}).\footnote{SafeVerify
(\weburl{github.com/GasStationManager/SafeVerify}) and Comparator
(\weburl{github.com/leanprover/comparator})
 are independent tools designed for the same problem: verifying that a potentially adversarial
submission proves the stated theorem.
SafeVerify predates Comparator, and was used by \citet{tsoukalas2026apn}.
As a cross-check, we re-verified all Claude Opus 4.8 submissions on the 492 \bench{} conjectures with Comparator.
It confirmed all SafeVerify-accepted solves except submissions to five
conjectures whose Lean formalizations had an unusual defect,
and it additionally verified two proofs that SafeVerify had rejected only
because the checker exhausted its resource limits. Under Comparator's
verdicts, Claude Opus 4.8 scores 144/492 (29\%) rather than 147/492 (30\%).
Future versions of this benchmark will use Comparator.}

Our agents have shell access to their own Lean environment, so they could in principle attack verification itself, for example by tampering with Mathlib, the Lean toolchain, or the statement file. We therefore split every attempt across three Docker containers, none of which has network access:

\begin{enumerate}
    \item The \emph{agent} container, where the model works on its proof.
    \item The \emph{compile} container, holding a clean Lean toolchain, where the submitted proof is compiled to a binary artifact (an olean).
    \item The \emph{scorer} container, which receives the submission olean, and runs SafeVerify on it.\footnote{More precisely, the scorer container compiles our own copy of the statement (trusted) into a reference olean, and receives the submission olean. It then runs SafeVerify on the pair.}
\end{enumerate}

This design defeats many classes of attacks:

\begin{enumerate}
    \item Tampering with the agent environment is powerless because only the submission's Lean source leaves the agent container.
    \item Malicious compile-time code. Lean elaboration can execute arbitrary code (e.g. a compile-time \texttt{\#eval}). Compilation to the binary olean is therefore confined in its own container, separate from the SafeVerify verdict.
    \item Proving a different statement. SafeVerify matches the submission against our own copy of the statement, requiring kernel-identical types.
    \item Redefining a definition the statement depends on, e.g. so that the conjecture becomes trivially true. SafeVerify requires the submission's definition bodies to be identical to the target's; only \texttt{sorry} stubs may be filled in.
    \item Smuggling in extra axioms. A declared axiom or a \texttt{sorry} (which introduces the axiom \texttt{sorryAx}) falls outside the three-axiom whitelist.
    \item Bypassing the kernel. Lean metaprogramming can insert declarations into an environment without kernel checking, and a buggy tactic can emit an ill-typed proof term; SafeVerify replays every submitted declaration through a fresh kernel. A \texttt{native\_decide} proof shifts trust from the kernel to the compiler, and is known to be subvertible via \texttt{@[implemented\_by]}; it introduces the axiom \texttt{Lean.ofReduceBool}, which falls outside the three-axiom whitelist.
\end{enumerate}

What remains trusted is Lean's kernel, SafeVerify itself, and the container isolation.

\subsection{AI agent}
\label{sec:agent}

Our base agent is deliberately simple: a ReAct-style tool loop built on the Inspect library \citep{inspectai}, with three tools. The tools are \texttt{bash}, a text editor, and a \texttt{resources} tool that reports remaining time and token budgets. The agent container provides a Lean 4 toolchain with Mathlib, the SageMath computer algebra system, and Python with \texttt{sympy}, \texttt{mpmath}, \texttt{numpy}, and \texttt{pantograph}.

Like in \citet{tsoukalas2026apn}, our agent can either prove or disprove the conjecture.\footnote{In \citet{tsoukalas2026apn}, the target theorem is an equivalence between a truth value and the conjecture, with the truth value left editable by the agent (see their Figure~1, where \texttt{EVOLVE-VALUE} markers are used to enclose expressions whose values the agent can change). Setting it to \texttt{True} and proving the equivalence proves the conjecture; setting it to \texttt{False} disproves it.} The agent iterates until it resolves the conjecture or hits a limit. Typically the binding limit is the per-conjecture spending cap (\$50 on the full set, \$200 on \textsc{Lite}), agents also cannot use more than 72 hours of working time.

Notably, all our agents are considerably simpler than the ``full-featured'' agent with which \citet{tsoukalas2026apn} obtained their OEIS results. Their agent runs an AlphaEvolve-style evolutionary search over a shared population of proof sketches: Lean files that compile but may leave subgoals unproven as \texttt{sorry} placeholders. Prover subagents (Gemini 3.1 Pro) mutate a sketch sampled from the population over a multi-turn episode with Lean compiler feedback. Subagents can call AlphaProof, a specialized system that searches the tree of Lean proof steps, on individual subgoals. AlphaProof's verdicts are memoized in a cache shared across the population. Since partial proofs have no objective fitness, rater subagents (Gemini 3.0 Flash) rank sketches on plausibility, clarity, and novelty; the rankings are aggregated into Elo scores, which bias the sampling of parent sketches via a P-UCB rule. The search ends when a sketch becomes \texttt{sorry}-free and passes validation.

\subsubsection{Literature}
\label{sec:agent-lit}

This variant gives agents access to an offline snapshot of the mathematics literature: the \LaTeX{} source trees of 476{,}000 pure-mathematics arXiv papers dated up to 2022.\footnote{The corpus is the arXiv subset of proof-pile \citep{proofpile}: papers in arXiv's mathematics archive (the \texttt{math} categories), as of proof-pile's compilation in 2022.}

\subsubsection{DeepAgent}
\label{sec:agent-deep}

The DeepAgent variant replaces the ReAct loop with Inspect's \texttt{deepagent}, which adds the following affordances: delegation to subagents, persistent memory, a todo-list tool, and a longer, opinionated system prompt.

\section{Results}

\subsection{AI performance}

\begin{figure}[h]
    \centering
    \includegraphics[width=\linewidth]{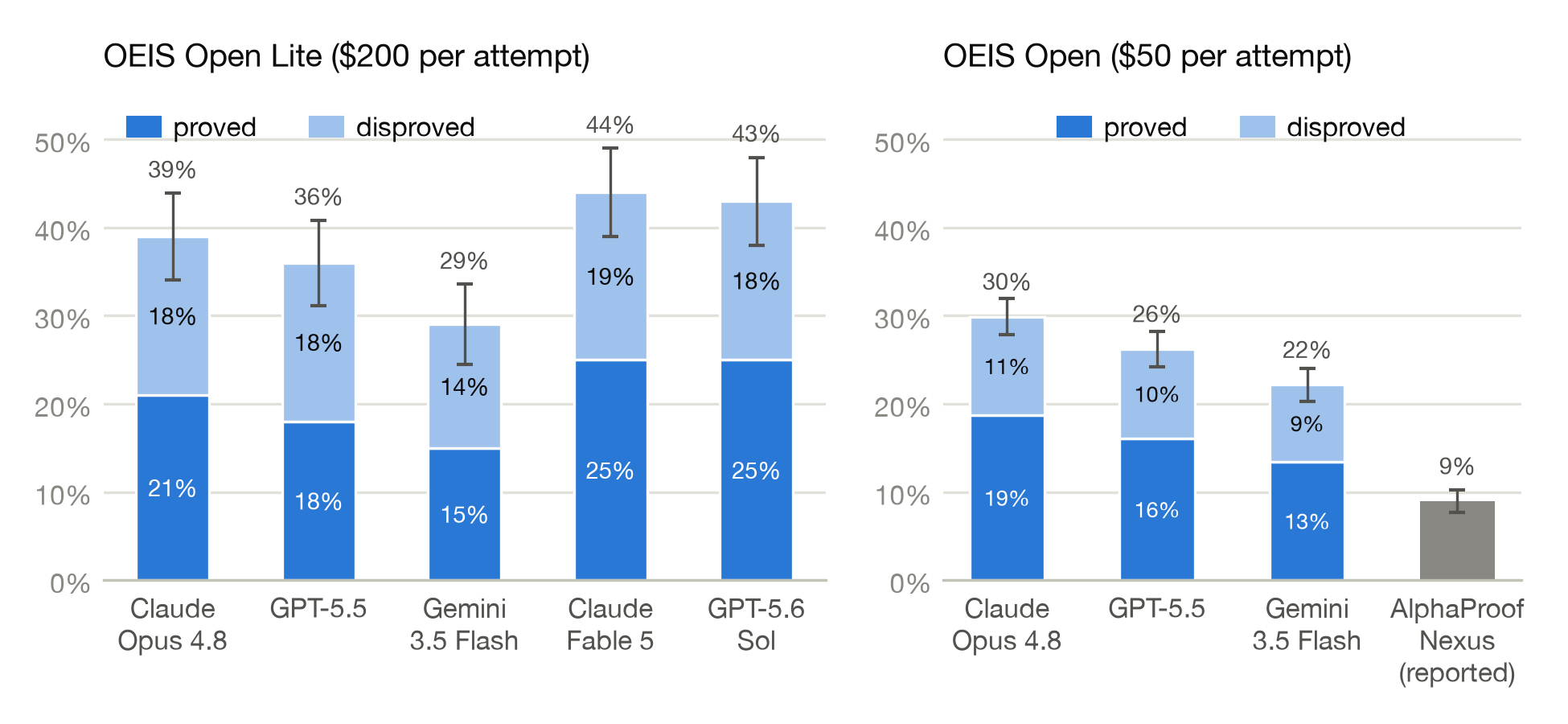}
    \caption{Accuracy on \benchlite{} with the base agent (left; 100 conjectures, \$200 budget per conjecture) and on the full \bench{} set, with the reported AlphaProof Nexus baseline (right; 492 conjectures, \$50 budget). Each bar splits solves into proofs (solid) and disproofs (pale). Error bars are $\pm$1 standard error.}
    \label{fig:accuracy}
\end{figure}

A model resolves a conjecture when it submits a proof of the conjecture or of its negation that passes verification. On the full \bench{} set, we ran each model once with a \$50 spending cap per conjecture.

\paragraph{Language models can resolve many open OEIS conjectures, and outperform AlphaProof Nexus.} Claude Opus 4.8 resolved 30\% of the 492 conjectures, GPT-5.5 resolved 26\%, and Gemini 3.5 Flash 22\% (Figure~\ref{fig:accuracy}, right). By comparison, \citet{tsoukalas2026apn} report that their system, AlphaProof Nexus, resolved 44 of the same 492 conjectures (9\%)\footnote{The paper reports 44/492, but the accompanying repository (\texttt{google-deepmind/alphaproof-nexus-results}) releases 38 OEIS proofs.}, at a similar cost to our result\footnote{Our average cost per resolved conjecture was \$6 (GPT-5.5), \$9 (Gemini 3.5 Flash), and \$10 (Claude Opus 4.8), with a maximum of \$47. In personal correspondence, the AlphaProof Nexus authors estimated that their cost per resolved OEIS conjecture was roughly \$10 on average, and up to about \$50 for the hardest few.}.

On \benchlite{}, with the cap raised to \$200 per conjecture, language models resolved 29\% (Gemini 3.5 Flash) to 44\% (Claude Fable 5) of the 100 conjectures (Figure~\ref{fig:accuracy}, left). Claude Fable 5 and GPT-5.6 Sol were evaluated only on \textsc{Lite}, and only with the base agent.

\paragraph*{Agent variants had no effect.} Neither giving models access to the mathematics literature, nor using the more complex DeepAgent (Sections~\ref{sec:agent-lit} and~\ref{sec:agent-deep}) affected the accuracy on \textsc{Lite} (Figure~\ref{fig:variants}, Appendix~\ref{sec:variant-results}).

\begin{figure}[htbp]
    \centering
    \includegraphics[width=\linewidth]{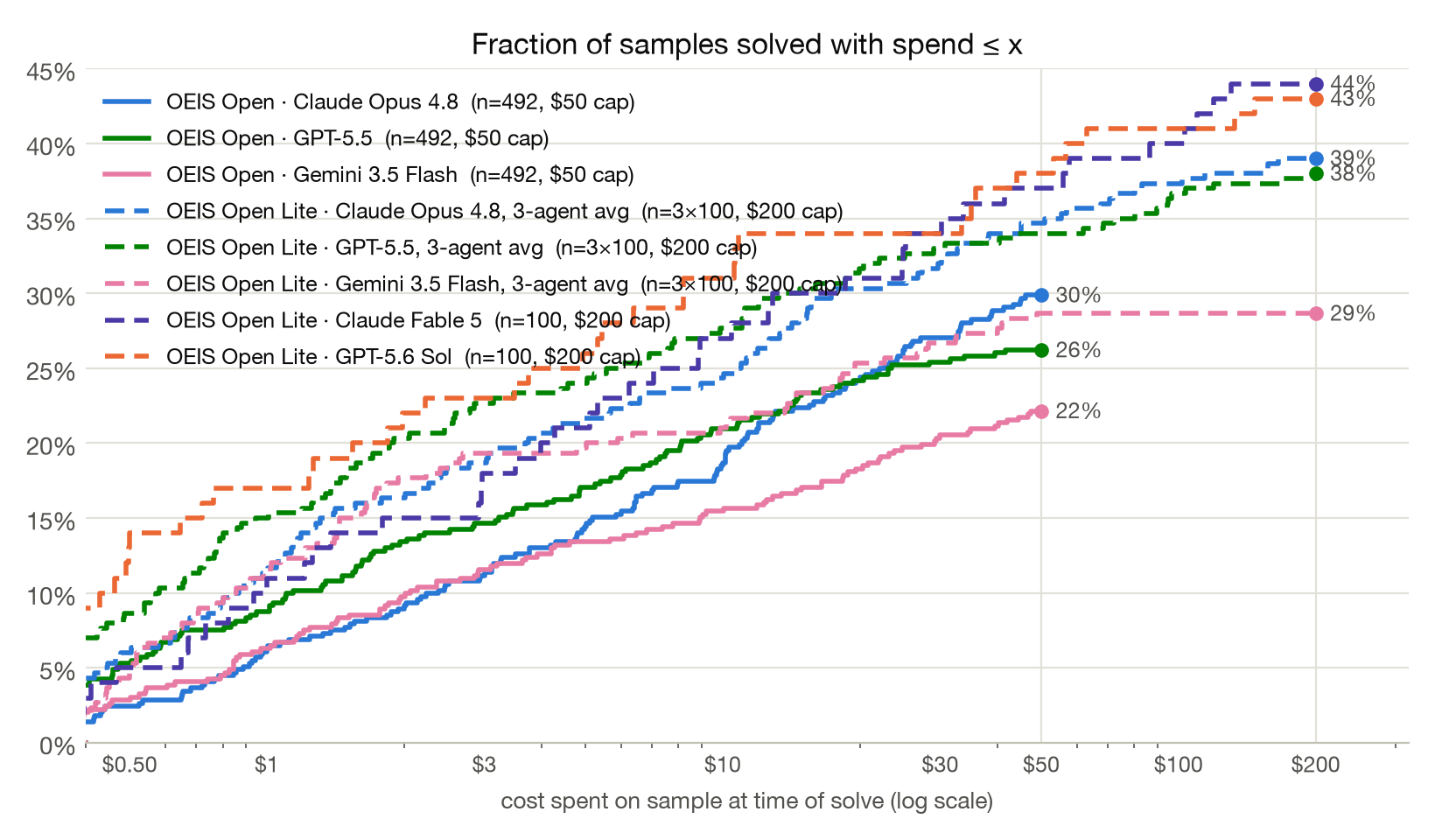}
    \caption{Solve rate as a function of per-sample spend: the fraction of conjectures resolved with spend at most $x$, where spend is measured at the moment the conjecture was resolved. Each curve ends at its run's per-conjecture budget cap. \textsc{Lite} curves average the three agent variants per model where available.}
    \label{fig:solve-cost-curves}
\end{figure}

\paragraph*{Solve rates appear to rise log-linearly with spend.} Figure~\ref{fig:solve-cost-curves} plots, for each run, the fraction of conjectures resolved as a function of the amount spent on that conjecture at the moment it was resolved. Reading a curve at $x$ estimates the solve rate of a run capped at $x$. The estimate is imperfect because agents are told their budget, which may affect their behavior. Solve rates rise roughly linearly in log-spend, on the order of ten percentage points per tenfold increase.

\FloatBarrier

\section{Discussion}

\paragraph*{Formalized open conjectures are reusable infrastructure.} This work builds on that of \citet{tsoukalas2026apn}, who collected and auto-formalized the conjectures, and released the statements openly on Google DeepMind's Formal Conjectures repository \citep{firsching2026formalconjectures}. We hope more datasets of this kind are released, and that this paper illustrates their value.

\paragraph*{A simple harness performed well.} Our base agent is minimal, yet it resolved more than three times as many conjectures as AlphaProof Nexus's elaborate evolutionary search.\footnote{AlphaProof Nexus's prover subagents use Gemini 3.1 Pro, released 19 February 2026; GPT-5.5 and Claude Opus 4.8 were released about two and three months later (23 April and 28 May 2026).} This fits the bitter lesson \citep{sutton2019bitter}: rather than prescribing how the model should work through problem-specific structure, it may be better to give a model simple tools and let it choose how to use them.

\paragraph*{On this benchmark, Claude Fable 5 and GPT-5.6 Sol do not represent a qualitative jump in autonomous AI proving.} The two newest-generation models scored highest on \benchlite{} (43--44\%), but their lead over the previous generation is only a few percentage points (Figure~\ref{fig:accuracy}, left).

\paragraph*{Further scaling inference would resolve more conjectures.} Because \benchlite{} is a random subset of \bench{}, its results estimate what a full-set run would achieve. At \$200 per conjecture, we would expect the best current models to resolve about 216 of the 492 conjectures, up from the 147 resolved at \$50. Nor is inference scaling exhausted at \$200: solve rates rose roughly log-linearly in spend without a clear plateau (Figure~\ref{fig:solve-cost-curves}), so larger budgets would likely push past 44\% (at exponential cost).

\subsection{Limitations}

\paragraph*{Though open, most conjectures have likely received little attention.} Current data does not let us reliably characterize the degree of mathematical interest these conjectures have received. However, it appears likely that most have received little attention. \citet{tsoukalas2026apn} started with a corpus of 2649 conjectures from the OEIS, open as of their work, and ``prompted Gemini to select 500 problems that are non-trivial, mathematically interesting, not famous open problems, and good candidates for automated theorem-proving''. Submissions to the OEIS must be approved by volunteer editors, who filter out ill-defined, contrived, and uninteresting sequences \citep{oeisContributionProcess, oeisWhatNotToSubmit}. Our metadata shows that, while there are 127 distinct proposers, the prolific conjecturer Zhi-Wei Sun proposed 37\% of \bench{} and 36\% of \benchlite{}. For 47\% of the conjectures, the underlying sequence's OEIS entry lists no links or references. Future work could use LMs to explicitly select problems that have received considerable mathematical interest rather than excluding them.

\paragraph*{Misformalization risk.} \citet{tsoukalas2026apn} explained that all 44 conjectures resolved by their system were reviewed by a human and no misformalizations were found. We have taken the Tsoukalas dataset as-is without performing further validation. It seems likely that at least some conjectures are misformalized. Misformalizations are likely to be easier to resolve than correctly formalized conjectures, and seem unlikely to strongly advantage one language model over another. Hence, despite the risk of misformalizations, \bench{} is still a useful tool for comparing language models. Future work could more carefully review whether formalizations were correct.

\paragraph*{No credit for reductions to famous open problems.} Our setup accepts only a proof of the conjecture or of its negation, but mathematicians also value results that relate a conjecture to a famous open problem. Proving that the conjecture implies a famous open problem, such as the Collatz conjecture, would often be considered the definitive word on it.

\paragraph*{Resolved conjectures may leak into training data.} Once a conjecture is resolved, its proof may enter the pretraining data of future models, which could then reproduce the proof rather than find it independently. The models evaluated in this paper have training cutoffs that predate the publication of \citet{tsoukalas2026apn}, so they cannot have learned the proofs released with that paper. For future models, the issue can be mitigated after the fact: conjectures resolved before a model's training cutoff can be filtered out, and all models compared on the remaining smaller set.

\bibliographystyle{unsrtnat}
\bibliography{references}

\clearpage
\appendix
\section{Appendix}

\subsection{Resolved conjectures}
\label{sec:top-solves}

Table~\ref{tab:top-solves} shows the 100 costliest of the 153 conjectures resolved by either Claude Fable 5 on \benchlite{} or Claude Opus 4.8 on the full set. The sequence descriptions, conjecture statements, and proof summaries were written by GPT-5.6 Sol agents given the accepted Lean proof, the sequence's OEIS entry, and sandboxed access to the Mathlib source tree. Cost is the total API spend of the run that resolved the conjecture; for the 35 conjectures resolved by both models we report the cheaper of the two. Each verdict links to the accepted Lean proof, which together with the rejected submissions and per-sample verifier output is available at \weburl{github.com/epoch-research/LeanOpenProblems-results}.

{\footnotesize
\renewcommand{\arraystretch}{1.15}
\begin{longtable}{p{0.42\textwidth} p{0.44\textwidth} r}
\caption{The 100 costliest conjectures resolved by either model, with LM-written summaries. Each verdict links to the accepted Lean proof.}
\label{tab:top-solves}\\
\hline
\textbf{Sequence and conjecture (LM-written)} & \textbf{Proof summary (LM-written)} & \textbf{Cost} \\ \hline
\endfirsthead
\multicolumn{3}{l}{\footnotesize\emph{Table \ref{tab:top-solves} (continued)}}\\
\hline
\textbf{Sequence and conjecture (LM-written)} & \textbf{Proof summary (LM-written)} & \textbf{Cost} \\ \hline
\endhead
\hline
\endfoot
\endlastfoot
\href{https://oeis.org/A364173}{A364173}: Conjecturally integral ratio \((9n)!\allowbreak(2n)!\allowbreak(3n/2)!/\allowbreak((9n/2)!\allowbreak(4n)!\allowbreak(3n)!n!)\)\par\smallskip \emph{Conjecture} (Peter Bala, 2023). For every prime \(p\ge 5\) and positive integers \(n,\allowbreak r\), \(a(np^r)\equiv a(np^{r-1})\pmod{p^{3r}}\). & \emph{\href{https://github.com/epoch-research/LeanOpenProblems-results/blob/main/runs/oeis-lite-200usd-fable-i7s7n9q7v4emgjp7/oeis_364173_conjecture_0/Submission/Spec.lean}{Proved.}} The proof first applies Legendre's Gamma duplication formula to rewrite the terms using ordinary factorials and odd-step products, then factors out powers of \(p\) and organizes the \(p\)-free factors into residue blocks. Reciprocal-sum cancellations, a Morley-type parity identity, and a quadratic \(p\)-adic product expansion make each shifted block \(p^{-3r}\)-close to \(1\); the assumed integrality and the \(p\)-adic norm--divisibility criterion then yield the congruence. & \$130 \\ \hline
\href{https://oeis.org/A333096}{A333096}: Degree-\(n\) Taylor sum of Catalan's generating function raised to \(4n\)\par\smallskip \emph{Conjecture} (Peter Bala, 2020). For every integer \(m\), prime \(p\ge 5\), and positive integers \(n,\allowbreak k\), the analogous sequence \(a_m\) obtained by replacing the exponent \(4n\) by \(mn\) satisfies \(a_m(np^k)\equiv a_m(np^{k-1})\pmod{p^{3k}}\). & \emph{\href{https://github.com/epoch-research/LeanOpenProblems-results/blob/main/runs/oeis-lite-200usd-fable-i7s7n9q7v4emgjp7/oeis_333096_supercongruence_conjecture/Submission/Spec.lean}{Proved.}} The proof rewrites \(a_m(N)\), up to a correction that cancels in differences, as \(\beta((m+1)N,\allowbreak N)=[X^N]G(X)(1-X)^{-(m+1)N}\), where \(G=(1-2X)/\allowbreak((1-X)(1-X+X^2))\). A Cartier fixed-point identity, the factorization of \(X^p+(1-X)^p-1\), antisymmetry cancellation, and Legendre--Kummer/Stirling-number valuation bounds yield \(p^{3(v+1)}\mid\beta(pS,\allowbreak pN)-\beta(S,\allowbreak N)\) whenever \(p^v\mid S,\allowbreak N\); taking \(v=k-1\) gives the result. & \$119 \\ \hline
\href{https://oeis.org/A333042}{A333042}: Coefficients of \(\exp\!\left(\sum_{k\ge1}\frac{(4k)!}{(k!)^4}\frac{x^k}{k}\right)\)\par\smallskip \emph{Conjecture} (Peter Bala, 2023). For every \(m\ge1\), let \(A_m\) be the displayed series with \(4\) replaced throughout by \(m\), and set \(b_m(n)=[x^n]A_m(x)^n\); then \(b_m(n)\) is integral and, for every prime \(p\ge5\) and \(n,\allowbreak r\ge1\), \(b_m(np^r)\equiv b_m(np^{r-1})\pmod{p^{3r}}\). & \emph{\href{https://github.com/epoch-research/LeanOpenProblems-results/blob/main/runs/oeis-lite-200usd-fable-i7s7n9q7v4emgjp7/general_supercongruence_conjecture/Submission/Spec.lean}{Proved.}} The proof works \(p\)-adically with formal power series: Dwork's integrality lemma and a Frobenius/correction-series factorization reduce the coefficient differences to valuation estimates supplied by a strengthened Kazandzidis binomial congruence and digit-carry bounds for tuples. The exceptional cases \(m=1,\allowbreak 2\) are evaluated explicitly as \(\tfrac12\binom{2N}{N}\) and \(\tfrac12\binom{4N}{N}\), respectively, after which the Kazandzidis congruence applies directly. & \$110 \\ \hline
\href{https://oeis.org/A357674}{A357674}: \(\binom{3n}{n}^4\allowbreak\left(\sum_{k=0}^{2n}\binom{n+k-1}{k}^2\right)^3\)\par\smallskip \emph{Conjecture} (Peter Bala, 2022). For every prime \(p\ge 3\), one has \(a(p)\equiv a(1)\pmod{p^5}\). & \emph{\href{https://github.com/epoch-research/LeanOpenProblems-results/blob/main/runs/oeis-lite-200usd-fable-i7s7n9q7v4emgjp7/A357674_conjecture_1/Submission/Spec.lean}{Proved.}} Working in \(\mathbb Q_p\), the proof uses the hockey-stick identity and product formulas for binomial coefficients, truncating the products in terms of harmonic and multiple harmonic sums; finite-field power-sum vanishing, \(k\mapsto p-k\) reflection, stuffle relations, and alternating-binomial/Vandermonde identities supply the required valuation bounds. The resulting expansions of the two factors modulo \(p^5\) have linear corrections that cancel after taking the fourth and third powers, while \(p=3,\allowbreak 5\) are checked directly. & \$103 \\ \hline
\href{https://oeis.org/A048153}{A048153}: \(\sum_{k=1}^{n} (k^2 \bmod n)\)\par\smallskip \emph{Conjecture} (Aspen A.M. Meissner, 2025). For every integer \(n\ge 1\), \(a(n)\le \left\lfloor (n^2-1)/2\right\rfloor\). & \emph{\href{https://github.com/epoch-research/LeanOpenProblems-results/blob/main/runs/oeis-lite-200usd-fable-i7s7n9q7v4emgjp7/oeis_48153_conjecture_0/Submission/Spec.lean}{Proved.}} Compare \(a(n)\) with the mirror sum \(a_-(n)=\sum_{k<n}(-k^2\bmod n)\): a finite cotangent-kernel identity and a divisor decomposition express \(a_-(n)-a(n)\) as a sum of nonnegative local terms. Their nonnegativity follows from explicit quadratic Gauss-sum evaluations (using the Jacobi theta functional equation, CRT/prime-power reduction, and Jacobi characters) and positivity of odd-character cotangent sums via Euler products for Dirichlet \(L\)-series and dominated convergence at \(s=1\); finally, \(a(n)+a_-(n)\le n(n-1)\) yields the bound. & \$86 \\ \hline
\href{https://oeis.org/A232616}{A232616}: Least \(m\): \(\{2^k-k:1\le k\le m\}\) covers residues mod \(n\).\par\smallskip \emph{Conjecture} (Zhi-Wei Sun, 2013). Writing \(p_n\) for the \(n\)-th prime, \(a(n)<2(p_n-1)\) for every \(n\ge 1\). & \emph{\href{https://github.com/epoch-research/LeanOpenProblems-results/blob/main/runs/oeis-lite-200usd-fable-i7s7n9q7v4emgjp7/oeis_232616_conjecture_i/Submission/Spec.lean}{Disproved.}} The counterexample is \(n=550172\): the period \(2^{k+29400}\equiv2^k\pmod{550172}\), a solved linear congruence, and a finite avoidance computation show that residue \(63945\) is not attained for \(k<17135927\), so \(a(550172)\ge17135927\). A bit-sieve gives \(p_{550172}<8567965\), hence \(2(p_{550172}-1)\le17135926\); a finite coverage certificate establishes nonemptiness of the defining set. & \$56 \\ \hline
\href{https://oeis.org/A381358}{A381358}: Row sums of irregular triangle A381587\par\smallskip \emph{Conjecture} (Paul D. Hanna, 2025). The limit \( \lim_{n\to\infty} a_n^{1/n}\) exists. & \emph{\href{https://github.com/epoch-research/LeanOpenProblems-results/blob/main/runs/oeis-full-50usd-ant-j0j0g4uzligm1k41/A381358_limit_exists/Submission/Spec.lean}{Proved.}} Stable prefix/suffix patterns under run-length encoding yield, for the row lengths \(\ell_n\) and the first and second iterated run counts \(r_n,\allowbreak q_n\), the eventual recurrences \(\ell_{n+1}=\ell_n+r_n\), \(r_{n+1}=r_n+q_n-1\), and \(q_{n+5}=q_{n+4}+\ell_n\), together with \(a_{n+1}=a_n+\ell_n\). For a \(\lambda\in(1,\allowbreak 2)\) satisfying \(\lambda^4(\lambda-1)^3=1\), a positive left-eigenvector functional obeys an affine first-order recurrence, giving \(a_n=\Theta(\lambda^n)\); squeezing the \(n\)-th roots then proves \(a_n^{1/n}\to\lambda\) (approximately \(1.5550575766\)). & \$46 \\ \hline
\href{https://oeis.org/A364176}{A364176}: Factorial ratio \((15n)!\allowbreak(5n/2)!\allowbreak(2n)!/\allowbreak((15n/2)!\allowbreak(6n)!\allowbreak(5n)!n!)\)\par\smallskip \emph{Conjecture} (Peter Bala, 2023). For every prime \(p\ge 5\) and positive integers \(n,\allowbreak r\), provided the two relevant terms are integral, \(a(np^r)\equiv a(np^{r-1})\pmod{p^{3r}}\). & \emph{\href{https://github.com/epoch-research/LeanOpenProblems-results/blob/main/runs/oeis-full-50usd-ant-j0j0g4uzligm1k41/oeis_364176_conjecture_0/Submission/Spec.lean}{Proved.}} Gamma recurrence first converts the fractional-factorial expression into an identity involving ordinary factorials, Gauss products of factors prime to \(p\), and a shifted product; splitting off all \(p\)-divisible factors then reduces the claim to block-product congruences. These congruences follow by pairing \(e\) with \(p^r-e\) and using the vanishing of the reciprocal-square sum over \((\mathbb Z/p^r\mathbb Z)^\times\), after which a factor coprime to \(p\) is canceled. & \$45 \\ \hline
\href{https://oeis.org/A365179}{A365179}: Prime powers \(p_n^6\) if \(p_n\equiv2\pmod3\), otherwise \(p_n^7\), except \(a(1)=2\).\par\smallskip \emph{Conjecture} (Jianing Song, 2023). For every \(n\ge2\), any finite group \(G\) with \(|\operatorname{Aut}(G)|=a(n)\) has \(|G|=a(n)/p_n\), and if \(p_n\equiv2\pmod3\), then such a group is unique up to isomorphism. & \emph{\href{https://github.com/epoch-research/LeanOpenProblems-results/blob/main/runs/oeis-full-50usd-ant-j0j0g4uzligm1k41/oeis_365179_conjecture_2/Submission/Spec.lean}{Disproved.}} An explicit group \(P\) on \((\mathbb Z/27\mathbb Z)^2\) is proved to have order \(729\) and exactly \(2187\) automorphisms by parametrizing automorphisms through generator images, using a Frattini-quotient generation criterion and a finite coordinate count. Power-order arguments then force automorphisms of \(C_2\times P\) to preserve both factors, so \(|\operatorname{Aut}(C_2\times P)|=2187=a(2)\), but \(|C_2\times P|=1458\ne2187/3\), disproving the order formula. & \$43 \\ \hline
\href{https://oeis.org/A226163}{A226163}: Determinants of Legendre-symbol matrices associated with the \(n\)-th prime\par\smallskip \emph{Conjecture} (Zhi-Wei Sun, 2013). For every \(n\ge 2\), \(a(n)=0\) if and only if \(p_n\equiv 3\pmod 4\). & \emph{\href{https://github.com/epoch-research/LeanOpenProblems-results/blob/main/runs/oeis-lite-200usd-fable-i7s7n9q7v4emgjp7/oeis_226163_conjecture_0/Submission/Spec.lean}{Proved.}} For \(p\equiv3\pmod4\), reindexing by nonzero quadratic residues rewrites the matrix, up to a nonzero scalar, as \(U-U^{\mathsf T}D\), where \(U\) is normal and \(D\) is a symmetric involution with determinant \(1\); an odd-dimensional determinant lemma proved by polynomial perturbation then forces singularity. For \(p\equiv1\pmod4\), Euler's criterion turns the entries modulo \(p\) into \((x_i+t_j)^{(p-1)/2}\), and a binomial--Vandermonde factorization together with Lagrange interpolation and Wilson's theorem proves that the determinant is nonzero. & \$42 \\ \hline
\href{https://oeis.org/A263001}{A263001}: Number of ordered positive pairs satisfying \(n=\pi(k(k+1))+\pi(m(m+1)/2)\)\par\smallskip \emph{Conjecture} (Zhi-Wei Sun, 2015). For every \(n>2\), \(a(n)>0\), and \(a(n)=1\) if and only if \(n\in\{1,\allowbreak 4,\allowbreak 6\}\). & \emph{\href{https://github.com/epoch-research/LeanOpenProblems-results/blob/main/runs/oeis-full-50usd-ant-j0j0g4uzligm1k41/oeis_a263001_conjecture/Submission/Spec.lean}{Disproved.}} A certified computation shows that \(a(51156)=1\), with unique pair \((k,\allowbreak m)=(16,\allowbreak 1119)\) (the two prime counts are \(58\) and \(51098\)), contradicting the claimed uniqueness classification. The formal disproof proves a packed-bit sieve correct, uses monotonicity to reduce to \(k\le 791\) and \(m\le 1119\), and evaluates the resulting finite count as a list convolution. & \$41 \\ \hline
\href{https://oeis.org/A261876}{A261876}: Ordered four-square representations with \(z>0\) and \(yz(5x^2+7y^2+9z^2)\) square\par\smallskip \emph{Conjecture} (Zhi-Wei Sun, 2016). For every positive integer \(n\), \(a(n)>0\), and \(a(n)=1\) if and only if \(n=4^k m\) for some \(k\ge 0\) and \(m\in\{1,\allowbreak 7,\allowbreak 23,\allowbreak 647,\allowbreak 863\}\). & \emph{\href{https://github.com/epoch-research/LeanOpenProblems-results/blob/main/runs/oeis-full-50usd-ant-j0j0g4uzligm1k41/oeis_261876_conjecture_0/Submission/Spec.lean}{Disproved.}} An exhaustive certificate bounds the coordinates by 263, uses base-\(2^{33}\) generating-function digit extraction to count the 34,692 four-square candidates at \(n=69{,\allowbreak }383\), and filters an explicit certified list to prove \(a(69{,\allowbreak }383)=1\), with sole survivor \((x,\allowbreak y,\allowbreak z,\allowbreak w)=(187,\allowbreak 147,\allowbreak 6,\allowbreak 113)\). But \(69{,\allowbreak }383\equiv3\pmod 4\) and is not one of the five proposed bases, so it cannot be \(4^k m\), disproving the claimed characterization. & \$39 \\ \hline
\href{https://oeis.org/A046969}{A046969}: Denominators in Stirling's expansion of \(\log\Gamma(z)\)\par\smallskip \emph{Conjecture} (Lorenzo Sauras Altuzarra, 2020). For every \(n\ge 2\) such that \(12\) divides \(a(n-1)\), \(a(n)\), and \(a(n+2)\), if \(a(n)/12\) is prime, then each of \(a(n-1)/12-(n-1)\), \(a(n)/12-n\), and \(a(n+2)/12-(n+2)\) is divisible by \(6\). & \emph{\href{https://github.com/epoch-research/LeanOpenProblems-results/blob/main/runs/oeis-full-50usd-ant-j0j0g4uzligm1k41/oeis_a046969_conjecture_2/Submission/Spec.lean}{Disproved.}} Using Faulhaber's identity to control \(p\)-adic valuations of \(B_{473582}\), together with certified modular power-sum computations and prime factorization, the disproof establishes at \(n=236791\) that \(a(n)=14172=12\cdot1181\), with \(1181\) prime, and verifies the required divisibility by \(12\) for the neighboring terms. But \(a(n)/12-n=1181-236791=-235610\) is not divisible by \(6\), giving a counterexample. & \$39 \\ \hline
\href{https://oeis.org/A141057}{A141057}: Abelian cubes of length \(3n\) over a three-letter alphabet\par\smallskip \emph{Conjecture} (Peter Bala, 2022). For every prime \(p\ge 5\) and positive integers \(n,\allowbreak k\), \(a(np^k)\equiv a(np^{k-1})\pmod{p^{3k}}\). & \emph{\href{https://github.com/epoch-research/LeanOpenProblems-results/blob/main/runs/oeis-full-50usd-ant-j0j0g4uzligm1k41/oeis_a141057_supercongruence_conjecture/Submission/Spec.lean}{Proved.}} The proof establishes the stronger one-step divisibility \(p^{3(v_p(m)+1)}\mid a(mp)-a(m)\) by splitting the multinomial-index lattice according to divisibility by \(p\): off-grid terms are controlled by Kummer-type binomial-valuation bounds, while on-grid terms are compared using a strengthened Jacobsthal--Kazandzidis-type binomial congruence. That congruence is obtained by removing multiples of \(p\) from factorial products, expanding the resulting shifted unit products, and using cancellation of reciprocal and reciprocal-square sums in \(\mathbb Z/p^s\mathbb Z\); taking \(m=np^{k-1}\) finishes the argument. & \$39 \\ \hline
\href{https://oeis.org/A361713}{A361713}: \(a(n)=\sum_{k=0}^{n-1}\binom{n}{k}^2\binom{n+k-1}{k}^2\)\par\smallskip \emph{Conjecture} (Peter Bala, 2023). For every prime \(p\ge 7\) and every \(r\ge 2\), \(a(p^r)\equiv a(p^{r-1})\pmod{p^{4r+1}}\). & \emph{\href{https://github.com/epoch-research/LeanOpenProblems-results/blob/main/runs/oeis-full-50usd-ant-j0j0g4uzligm1k41/oeis_361713_conjecture_2/Submission/Spec.lean}{Disproved.}} A certified modular recurrence tracks the \(7\)-adic valuations and unit parts of the binomial factors, using Newton-lifted inverses and chunked computation modulo \(7^{25}\). For \(p=7,\allowbreak r=6\) it yields \(a(7^6)-a(7^5)\equiv-7^{24}\pmod{7^{25}}\), so the difference has \(7\)-adic valuation \(24<25\), disproving the conjecture. & \$35 \\ \hline
\href{https://oeis.org/A361883}{A361883}: \(\frac1n\sum_{k=0}^n(n+2k)\binom{n+k-1}{k}^3\)\par\smallskip \emph{Conjecture} (Peter Bala, 2023). For every prime \(p\ge 5\) and all positive integers \(n,\allowbreak r\), one has \(a(np^r)\equiv a(np^{r-1})\pmod{p^{3r}}\). & \emph{\href{https://github.com/epoch-research/LeanOpenProblems-results/blob/main/runs/oeis-lite-200usd-fable-i7s7n9q7v4emgjp7/oeis_361883_conjecture_0/Submission/Spec.lean}{Proved.}} Working in \(\mathbb Q_p\), the proof factors the relevant binomial coefficients using \(E(N,\allowbreak k)=\prod_{1\le i\le k,\allowbreak \,\allowbreak p\nmid i}(1+N/i)\) and splits the sum according to \(p\mid k\); a second-order expansion of \(E\), Wolstenholme-type bounds for reciprocal sums, and a block-descent for auxiliary inverse-square sums bound both parts by \(p^{-3r}\). A binomial identity establishes integrality, after which the norm estimate gives divisibility by \(p^{3r}\), and extracting the \(p\)-part of \(n\) removes the intermediate assumption \(p\nmid n\). & \$34 \\ \hline
\href{https://oeis.org/A002897}{A002897}: Cubes of central binomial coefficients: $\binom{2n}{n}^3$\par\smallskip \emph{Conjecture} (Peter Bala, 2022). For every $n\ge 0$, $a(n)=[x^n y^n z^n](1+x+y+z)^{2n}(1+x+y-z)^n(1+x-y+z)^n$. & \emph{\href{https://github.com/epoch-research/LeanOpenProblems-results/blob/main/runs/oeis-full-50usd-ant-j0j0g4uzligm1k41/oeis_2897_conjecture_0/Submission/Spec.lean}{Proved.}} The proof realizes extraction in $y,\allowbreak z$ through iterated commuting partial derivatives and uses the sign symmetries of the three linear factors to reduce the target coefficient, after setting $y=z=0$, to a binomial sum. Factoring the paired forms as $(1+x)^2-(y-z)^2$ isolates the surviving terms, and Vandermonde's identity $\sum_{k=0}^n\binom nk^2=\binom{2n}{n}$, together with factorial cancellation, completes the evaluation. & \$33 \\ \hline
\href{https://oeis.org/A251758}{A251758}: Floor of \(n^2/\sum_i d_i d_{i+1}\) for ordered divisors\par\smallskip \emph{Conjecture} (Robert G. Wilson v, 2014). For \(k=1,\allowbreak 2,\allowbreak 3,\allowbreak 4,\allowbreak 5,\allowbreak 6,\allowbreak 7,\allowbreak 9,\allowbreak 10,\allowbreak 11,\allowbreak 12,\allowbreak 13,\allowbreak 14,\allowbreak 15,\allowbreak 16,\allowbreak 17\), the least \(x\ge2\) with \(a(x)=k\) is, respectively, \(4,\allowbreak 2,\allowbreak 3,\allowbreak 25,\allowbreak 5,\allowbreak 49,\allowbreak 7,\allowbreak 2431,\allowbreak 121,\allowbreak 11,\allowbreak 169,\allowbreak 13,\allowbreak 6678671,\allowbreak 7429,\allowbreak 289,\allowbreak 17\). & \emph{\href{https://github.com/epoch-research/LeanOpenProblems-results/blob/main/runs/oeis-full-50usd-ant-j0j0g4uzligm1k41/a251758_conjecture_first_occurrences/Submission/Spec.lean}{Proved.}} The proof establishes the correctness of an efficient divisor enumeration pairing divisors up to \(\sqrt n\) with their complements, then verifies the candidates and, except for \(k=14\), exhaustively excludes smaller indices by direct computation. For \(k=14\), divisor reflection and a termwise AM--GM bound reduce the problem to an Euler-product estimate for \(\sigma_2(n)/n^2\); the least prime factor is at least \(17\), and below \(17\cdot19\cdot23\cdot29\cdot31=6678671\) there are at most four distinct prime factors, yielding a bound incompatible with \(a(n)=14\). & \$33 \\ \hline
\href{https://oeis.org/A281267}{A281267}: $[x^n]\prod_{k=1}^n(1-x^k)^{nk}$\par\smallskip \emph{Conjecture} (Peter Bala, 2023). For every prime \(p\ge 3\) and all positive integers \(n,\allowbreak k\), \(a(np^k)\equiv a(np^{k-1})\pmod{p^{2k}}\). & \emph{\href{https://github.com/epoch-research/LeanOpenProblems-results/blob/main/runs/oeis-full-50usd-ant-j0j0g4uzligm1k41/oeis_281267_conjecture_0/Submission/Spec.lean}{Proved.}} A universal power series \(F\) is introduced with \(a(M)=[x^M]F^M\); its logarithmic derivative \(xF'/F=-\sum_{r\ge1}\sigma_2(r)x^r\) supplies coefficient recurrences, while finite polynomial approximations establish integrality. For the Dwork quotient \(\Phi_p=F(x)^p/F(x^p)\), divisor-sum identities, the expansion \(\Phi_p^M=\exp(M\Psi_p)\), and ultrametric \(p\)-adic coefficient and factorial bounds prove the stronger congruence \(a(Mp)\equiv a(M)\pmod{p^{2v_p(M)+2}}\), from which the claim follows by taking \(M=np^{k-1}\). & \$33 \\ \hline
\href{https://oeis.org/A052709}{A052709}: Coefficients of \((1-\sqrt{1-4x-4x^2})/\allowbreak(2(1+x))\).\par\smallskip \emph{Conjecture} (Gus Wiseman, 2021). For every \(n>0\), \(a(n)\) is the number of length-\(n-1\) sequences of positive integers whose set of values is \(\{1,\allowbreak \ldots,\allowbreak m\}\) for some \(m\ge 0\), and which have no indices \(i<j<k\) with \(x_i\le x_j\le x_k\). & \emph{\href{https://github.com/epoch-research/LeanOpenProblems-results/blob/main/runs/oeis-full-50usd-ant-j0j0g4uzligm1k41/oeis_52709_conjecture_0/Submission/Spec.lean}{Proved.}} The proof partitions the sequences by their maximum \(M\) and excess length \(s=(n-1)-M\), refining further by the length of the initial strictly decreasing prefix. Avoidance forces \(M\) to occur once or twice, so deleting and reinserting its copies gives bijective recurrences whose ballot-number solution yields \(C_M\binom{M}{s}\); summing over \(M\) gives the Catalan--binomial formula for \(a(n)\). & \$33 \\ \hline
\href{https://oeis.org/A051293}{A051293}: Nonempty subsets of \(\{1,\allowbreak \ldots,\allowbreak n\}\) with integer average\par\smallskip \emph{Conjecture} (Benoit Cloitre, 2003). Writing \(a(n)\) for the sequence, as \(n\to\infty\), \(a(n)=\frac{2^{n+1}}{n}\left(1+\frac1n+\frac3{n^2}+\frac{13}{n^3}+\frac{75}{n^4}+\frac{541}{n^5}\right)+o\!\left(\frac{2^{n+1}}{n^6}\right).\) & \emph{\href{https://github.com/epoch-research/LeanOpenProblems-results/blob/main/runs/oeis-full-50usd-ant-j0j0g4uzligm1k41/oeis_51293_conjecture_0/Submission/Spec.lean}{Proved.}} Grouping by subset size and applying a roots-of-unity filter reduces the count to \(M_n=\sum_{k=1}^n\binom nk/k=\sum_{j=1}^n(2^j-1)/j\), up to an exponentially smaller error; the latter is controlled by factoring the generating polynomial into root-of-unity blocks and bounding coefficient \(\ell^1\)-norms. Reversing the sum for \(M_n\), expanding \(1/(n-i)\), and evaluating the geometric moments \(\sum_{i\ge0}i^t2^{-i}\) yields the six coefficients, while explicit tail bounds give the stated little-\(o\) remainder. & \$33 \\ \hline
\href{https://oeis.org/A357569}{A357569}: \(\binom{3n}{n}^2-27\binom{2n}{n}\)\par\smallskip \emph{Conjecture} (Peter Bala, 2022). For every prime \(p\ge 3\) and every integer \(r\ge 2\), one has \(a(p^r)\equiv a(p^{r-1})\pmod{p^{3r+3}}\). & \emph{\href{https://github.com/epoch-research/LeanOpenProblems-results/blob/main/runs/oeis-lite-200usd-fable-i7s7n9q7v4emgjp7/oeis_357569_conjecture_0/Submission/Spec.lean}{Proved.}} Working in \(\mathbb Q_p\), the proof expresses the ratios of the successive binomial coefficients as unit products \(\beta_t=\prod_{p\nmid k}(1+t p^r/k)\), and uses reflection and doubling on the units modulo \(p^r\) to obtain sharp valuation bounds for the reciprocal power sums. A fourth-degree Newton expansion of these products, together with the key cancellation between its \(t=2\) and \(t=1\) instances and auxiliary Wolstenholme-type bounds, gives valuation at least \(3r+3\), hence the required divisibility. & \$30 \\ \hline
\href{https://oeis.org/A181546}{A181546}: $a(n)=\sum_{k=0}^{\lfloor n/2\rfloor}\binom{n-k}{k}^4$\par\smallskip \emph{Conjecture} (Paul D. Hanna, 2010). For the analogous sums $F_L(n)$ obtained by replacing the exponent $4$ by any integer $L\ge 0$, one has $ \lim_{n\to\infty}\frac{F_L(n+1)}{F_L(n)}=\frac{\operatorname{Fib}(L)\sqrt5+\operatorname{Lucas}(L)}2$. & \emph{\href{https://github.com/epoch-research/LeanOpenProblems-results/blob/main/runs/oeis-full-50usd-ant-j0j0g4uzligm1k41/oeis_181546_conjecture_0/Submission/Spec.lean}{Proved.}} Writing $b_{n,\allowbreak k}=\binom{n-k}{k}$, the proof uses exact adjacent-term ratios to establish geometric concentration around the saddle point $k\sim(5-\sqrt5)n/10$; on the central window, $b_{n+1,\allowbreak k}/b_{n,\allowbreak k}=(n+1-k)/\allowbreak(n+1-2k)\to\varphi$, while polynomial-times-geometric bounds make the tails negligible (with $L=0$ handled directly). Binet identities then identify $\varphi^L$ with the stated Fibonacci--Lucas expression. & \$27 \\ \hline
\href{https://oeis.org/A309391}{A309391}: \(\gcd(n,\allowbreak \operatorname{num}(H_{n-2})-\operatorname{den}(H_{n-2}))\) for \(n>2\)\par\smallskip \emph{Conjecture} (Amiram Eldar and Thomas Ordowski, 2019). For \(n>2\), \(A309391(n)=n\) implies that \(n\) is prime. & \emph{\href{https://github.com/epoch-research/LeanOpenProblems-results/blob/main/runs/oeis-full-50usd-ant-j0j0g4uzligm1k41/A309391_conjecture/Submission/Spec.lean}{Disproved.}} Let \(p=16843\), a Wolstenholme prime: a certified modular-inverse sum modulo \(p^3\) proves \(v_p(H_{p-1})\ge 3\). Splitting the relevant harmonic sum into terms divisible and not divisible by \(p\), and pairing the latter by \(j\leftrightarrow p^2-j\), shows \(p^2\mid \operatorname{num}(H_{p^2-2})-\operatorname{den}(H_{p^2-2})\); hence \(n=p^2=283686649\) is composite but \(A309391(n)=n\). & \$26 \\ \hline
\href{https://oeis.org/A372761}{A372761}: Denominators of \(1/(2-3/(3-4/(\cdots-(n-1)-n/(n+4)))))\)\par\smallskip \emph{Conjecture} (Mohammed Bouras, 2024). Every odd prime \(p\notin\{3,\allowbreak 5\}\) occurs as a term at exactly one index \(n\ge 3\). & \emph{\href{https://github.com/epoch-research/LeanOpenProblems-results/blob/main/runs/oeis-full-50usd-ant-j0j0g4uzligm1k41/oeis_372761_conjecture_2/Submission/Spec.lean}{Proved.}} Introducing a continuant recurrence \(B_{n,\allowbreak k}\), a conserved invariant gives \(a(n)=(5n-4)/\gcd(|B_{n,\allowbreak 3}|,\allowbreak 5n-4)\), while telescoping the recurrence yields factorial divisibility identities controlling this gcd. For each \(p\ge7\), one takes the representative \(3\le n_0<p\) satisfying \(5n_0\equiv4\pmod p\); the identities prove \(a(n_0)=p\) (with \(p=11,\allowbreak n_0=3\) checked directly), and a \(p\)-adic valuation/factorial bound rules out occurrences with \(n\ge p\), leaving uniqueness modulo \(p\). & \$25 \\ \hline
\href{https://oeis.org/A263326}{A263326}: Denominator of $\sum_{d\mid n} 1/(d+1)$\par\smallskip \emph{Conjecture} (Zhi-Wei Sun, 2015). For every pair of positive integers \(k,\allowbreak s\), the numbers \(\sum_{d\mid n}1/(d+k)^s\), as \(n\) ranges over the positive integers, have pairwise distinct fractional parts and are never integers. & \emph{\href{https://github.com/epoch-research/LeanOpenProblems-results/blob/main/runs/oeis-full-50usd-ant-j0j0g4uzligm1k41/oeis_a263326_conjecture_1/Submission/Spec.lean}{Disproved.}} An explicit counterexample takes \(k=31\), \(s=1\), and the distinct indices \(5\) and \(1829=31\cdot59\): enumerating their divisors and simplifying exactly gives both sums as \(17/288\). Equivalently, the key unit-fraction identity is \(1/62+1/90+1/1860=1/36\), so the two values---and hence their fractional parts---coincide. & \$25 \\ \hline
\href{https://oeis.org/A340079}{A340079}: \(n/\gcd(n,\allowbreak 1+\sum_{k=1}^n \gcd(k,\allowbreak n))\)\par\smallskip \emph{Conjecture} (Thomas Ordowski, 2014). \(a(n)=1\) if and only if \(n=1\) or \(n\) is prime. & \emph{\href{https://github.com/epoch-research/LeanOpenProblems-results/blob/main/runs/oeis-lite-200usd-fable-i7s7n9q7v4emgjp7/oeis_340079_conjecture_0/Submission/Spec.lean}{Disproved.}} A counterexample is \(n=23492890653051=3\cdot37\cdot43\cdot42307\cdot116341\). Writing the gcd-sum as Pillai's multiplicative function \(P(n)=\sum_{d\mid n}\varphi(d)(n/d)\), with \(P(p)=2p-1\) for primes, gives \(P(n)=610815156979325=26n-1\); thus \(n\mid 1+P(n)\), so \(a(n)=1\) although \(n\) is composite. & \$25 \\ \hline
\href{https://oeis.org/A354747}{A354747}: Least \(m\ge1\) such that \(2n\cdot3^m-1\) is prime\par\smallskip \emph{Conjecture} (Felix Fr\"{o}hlich, 2022). \(a(100943)=0\). & \emph{\href{https://github.com/epoch-research/LeanOpenProblems-results/blob/main/runs/oeis-lite-200usd-fable-i7s7n9q7v4emgjp7/oeis_a354747_conjecture_0/Submission/Spec.lean}{Disproved.}} At \(m=39101\), the 18,662-digit number \(N=2\cdot100943\cdot3^{39101}-1\) is prime, as certified by an \(N+1\) Lucas test in \((\mathbb Z/p\mathbb Z)[\sqrt3]\): segmented binary exponentiation and \(N+1=2\cdot100943\cdot3^{39101}\) show that \(2+\sqrt3\) has exact order \(N+1\) for every prime \(p\mid N\). The bound on the quadratic ring's unit group then forces \(p^2>N\) for every prime divisor \(p\) of \(N\), hence \(N\) is prime and \(a(100943)\ne0\). & \$25 \\ \hline
\href{https://oeis.org/A120424}{A120424}: Half-Fibonacci: add previous two terms, halving each if even\par\smallskip \emph{Conjecture} (Reed Kelly, 2006). The even-valued terms have natural density \(1/2\), and infinitely many adjacent pairs of terms differ by \(1\). & \emph{\href{https://github.com/epoch-research/LeanOpenProblems-results/blob/main/runs/oeis-full-50usd-ant-j0j0g4uzligm1k41/oeis_a120424_conjecture_0/Submission/Spec.lean}{Disproved.}} Assuming density \(1/2\), the even-term counts in long prefixes are eventually trapped between \(49.5\%\) and \(50.5\%\); a growth monovariant \(P_k=a_k+2a_{k+1}\), together with a count of steps whose endpoints are not both even, supplies exponential lower bounds on the terms. A backward-doubling argument for the first differences of the halved-value transform then shows that any sufficiently late difference-\(1\) pair forces an all-even block of length about \(n/36\), contradicting those density bounds, so the two claims cannot both hold. & \$24 \\ \hline
\href{https://oeis.org/A234360}{A234360}: Count of \(0<k<n\) such that \((k+1)^{\phi(n-k)}+k\) is prime\par\smallskip \emph{Conjecture} (Zhi-Wei Sun, 2013). For every \(n>1\), \(a(n)>0\), and for every \(n>5\) there exists \(0<k<n\) such that \((k+1)^{\phi(n-k)/2}-k\) is prime. & \emph{\href{https://github.com/epoch-research/LeanOpenProblems-results/blob/main/runs/oeis-full-50usd-ant-j0j0g4uzligm1k41/oeis_234360_conjecture_0/Submission/Spec.lean}{Disproved.}} The second assertion fails at \(n=1408\): an exhaustive case split over \(1\le k<1408\) proves that every candidate \((k+1)^{\phi(1408-k)/2}-k\) is nonprime. After computing the totients, the proof uses explicit proper-divisor certificates in most cases, base-2 Fermat-congruence witnesses for the remaining large cases, and direct evaluation for the final small values. & \$24 \\ \hline
\href{https://oeis.org/A228143}{A228143}: Hankel determinants of the Ap\'{e}ry numbers A005259\par\smallskip \emph{Conjecture} (Peter Bala, 2018). If \(A(x)=\sum_{n\ge0}a(n)x^n\), then \(A(x/3)^{1/8}\in\mathbb Z[[x]]\) (equivalently, \(A(x/3)\) is an eighth power in \(\mathbb Z[[x]]\)). & \emph{\href{https://github.com/epoch-research/LeanOpenProblems-results/blob/main/runs/oeis-full-50usd-ant-j0j0g4uzligm1k41/oeis_228143_conjecture_1/Submission/Spec.lean}{Proved.}} Lucas's theorem modulo \(3\) and the evenness of central binomial coefficients give the entry congruences \(A_n\equiv(-1)^n\pmod 3\) and \(A_n\equiv1\pmod 4\); unitriangular column reductions then yield \(3^n\mid a(n)\) and \(4^n\mid a(n)\), with \(16\mid a(n)\) for \(n>0\) using \(a(1)=48\). Substituting \(A(x/3)-1\) into the binomial series \((1+X)^{1/8}\), prime-by-prime \(p\)-adic estimates---especially cancellation from these divisibilities at \(p=2,\allowbreak 3\)---show that every coefficient has denominator one, producing the required integral eighth root. & \$23 \\ \hline
\href{https://oeis.org/A309132}{A309132}: Denominator of \(\operatorname{num}(B_{n-1})/n+\operatorname{den}(B_{n-1})/n^2\)\par\smallskip \emph{Conjecture} (Thomas Ordowski, 2019). The assertion that, for every \(n>1\), \(a(n)=1\) if and only if \(n\) is prime is equivalent to the Agoh--Giuga conjecture that \(n\) is prime if and only if \(\sum_{k=1}^{n-1} k^{n-1}\equiv-1\pmod n\). & \emph{\href{https://github.com/epoch-research/LeanOpenProblems-results/blob/main/runs/oeis-full-50usd-ant-j0j0g4uzligm1k41/oeis_309132_conjecture_2/Submission/Spec.lean}{Proved.}} Writing \(B_{n-1}=N/D\), the proof reduces \(a(n)=1\) to \(n^2\mid Nn+D\), then proves this equivalent to \(n\mid \sum_{k=0}^{n-1}k^{n-1}+1\). The prime-power-local argument combines Faulhaber's formula and \(p\)-adic valuations (yielding the von Staudt--Clausen denominator criterion) with the finite-field power-sum identity over \(\mathbb F_p\), handling even \(n\) separately. & \$23 \\ \hline
\href{https://oeis.org/A211420}{A211420}: Factorial ratio $(8n)!n!/\allowbreak((4n)!\allowbreak(3n)!\allowbreak(2n)!)$\par\smallskip \emph{Conjecture} (Peter Bala, 2025). For every \(r\ge 0\), there is a positive integer \(K(r)\), independent of \(n\), such that \(\prod_{k=0}^{r-1}(8n-(2k+1))\mid K(r)a(n)\) for every \(n\ge 0\). & \emph{\href{https://github.com/epoch-research/LeanOpenProblems-results/blob/main/runs/oeis-full-50usd-ant-j0j0g4uzligm1k41/oeis_211420_conjecture_1/Submission/Spec.lean}{Proved.}} The proof gives the explicit uniform choice \(K(r)=(8r)!\); when \(r\le 4n\), Legendre's formula converts a crucial auxiliary factorial divisibility into a floor inequality at each prime power (proved by reduction modulo that power), and a telescoping factorial identity for the odd-factor product then yields the claim. When \(r>4n\), taking absolute values splits the product into two odd double-factorials, whose standard factorial formulas show directly that it divides \((8r)!\). & \$22 \\ \hline
\href{https://oeis.org/A114362}{A114362}: Numerators of \(\zeta(4n)/\zeta(2n)^2\), with \(a(0)=2\)\par\smallskip \emph{Conjecture} (Thomas Ordowski, 2022). Writing \(t(n)=\zeta(2n)/\zeta(n)^2\), one has \( \frac{1-t(n)}{1+t(n)}=2^{-n}+3^{-n}+5^{-n}+7^{-n}+O(11^{-n})\) as \(n\to\infty\). & \emph{\href{https://github.com/epoch-research/LeanOpenProblems-results/blob/main/runs/oeis-full-50usd-ant-j0j0g4uzligm1k41/oeis_A114362_conjecture_1/Submission/Spec.lean}{Proved.}} The proof rewrites \(\zeta(n)\) and \(\zeta(2n)\) as Dirichlet series, splits off their initial terms, and uses exact polynomial cancellation to show that every remaining finite product decays at least as fast as \(11^{-n}\). Comparison of the infinite tails with the convergent square series then gives the explicit bound \(725\cdot 11^{-n}\) for all \(n\ge4\), establishing the Big-O estimate. & \$20 \\ \hline
\href{https://oeis.org/A227582}{A227582}: \(a(n)=\left\lfloor(6n^2+6n-1)/5\right\rfloor\).\par\smallskip \emph{Conjecture} (Clark Kimberling, 2025). For every integer \(n\ge 1\), \(a(n)=\left\lfloor\bigl(2H_n-H_{n^2+n-1}-\gamma\bigr)^{-1}\right\rfloor\), where \(H_n\) is the \(n\)-th harmonic number and \(\gamma\) is the Euler--Mascheroni constant. & \emph{\href{https://github.com/epoch-research/LeanOpenProblems-results/blob/main/runs/oeis-full-50usd-ant-j0j0g4uzligm1k41/oeis_227582_conjecture_0/Submission/Spec.lean}{Proved.}} Strong induction reduces the order-7 recurrence to a quadratic closed form with a period-5 correction. Monotone Euler--Maclaurin-type corrections to \(H_m-\log m\) converging to \(\gamma\), together with truncated \(\operatorname{artanh}\)-series bounds for \(\log((n^2+n-1)/n^2)\), sandwich the denominator tightly enough that residue-class rational inequalities give \(a(n)\le (2H_n-H_{n^2+n-1}-\gamma)^{-1}<a(n)+1\), determining the floor. & \$19 \\ \hline
\href{https://oeis.org/A064169}{A064169}: Numerator minus denominator of the n-th harmonic number\par\smallskip \emph{Conjecture} (Amiram Eldar and Thomas Ordowski, 2019). For every natural number \(n>2\), \(n\mid a(n-2)\) if and only if \(n\) is prime. & \emph{\href{https://github.com/epoch-research/LeanOpenProblems-results/blob/main/runs/oeis-full-50usd-ant-j0j0g4uzligm1k41/oeis_64169_conjecture_0/Submission/Spec.lean}{Disproved.}} For \(p=16843\), a certified modular recurrence modulo \(p^3\) establishes the Wolstenholme-prime property \(v_p(H_{p-1})\ge 3\); splitting \(H_{p^2-2}\) into indices divisible and not divisible by \(p\), then pairing \(k\) with \(p^2-k\), yields \(v_p(H_{p^2-2}-1)\ge 2\). Hence \(p^2\mid a(p^2-2)\), although \(p^2=283686649\) is composite. & \$19 \\ \hline
\href{https://oeis.org/A300997}{A300997}: Stabilization time of rightward floor-halving automaton from mass \(n\)\par\smallskip \emph{Conjecture} (Luc Rousseau, 2018). For every \(n\ge 1\), \(a(n+1)=a(n)+1\) or \(a(n+1)=a(n)+2\). & \emph{\href{https://github.com/epoch-research/LeanOpenProblems-results/blob/main/runs/oeis-lite-200usd-fable-i7s7n9q7v4emgjp7/oeis_a300997_finite_difference_is_one_or_two/Submission/Spec.lean}{Proved.}} The proof recasts the list evolution as an equivalent function-valued, mass-conserving, hole-free process and couples the systems of masses \(n\) and \(n+1\) by tracking their single extra grain, which moves right exactly when the mass beneath it is odd. A nearby-nonunit invariant and the characterization \([1^{n-2},\allowbreak 2]\) of the unique nonstable predecessor of the target show that this grain is one or two sites short of position \(n\) when the mass-\(n\) system first settles; it then advances one site per step, so the coupled system settles exactly one or two steps later. & \$19 \\ \hline
\href{https://oeis.org/A248802}{A248802}: Smallest prime factor of \(2^{2^n+2}+3\)\par\smallskip \emph{Conjecture} (Chai Wah Wu, 2019). For every \(n\ge 0\), \(a(58n+26)=1399\) provided \(58n+26\) belongs to none of the families \(10m+2\), \(36m+16\) with \(m\not\equiv1\pmod 5\), and \(84m+22\) with \(m\not\equiv0\pmod 5\) (\(m\ge0\)). & \emph{\href{https://github.com/epoch-research/LeanOpenProblems-results/blob/main/runs/oeis-full-50usd-ant-j0j0g4uzligm1k41/oeis_248802_conjecture_4/Submission/Spec.lean}{Proved.}} Reducing exponents modulo \(233\) shows that \(2^{2^{58n+26}+2}\equiv-3\pmod{1399}\), so \(1399\) always divides the relevant term. To prove minimality, multiplicative-order periodicity reduces each prime \(p<1399\) to finitely many residues of \(n\), checked by a proved-correct fast binary modular-exponentiation routine; the noncoverage assumptions eliminate the exceptional residues for \(p=67,\allowbreak 271,\allowbreak 523\). & \$18 \\ \hline
\href{https://oeis.org/A317940}{A317940}: Numerators of the Dirichlet convolution square root of A046644\par\smallskip \emph{Conjecture} (Antti Karttunen, 2018). For every integer \(n\ge 1\), the underlying rational coefficient \(f(n)\) is nonnegative. & \emph{\href{https://github.com/epoch-research/LeanOpenProblems-results/blob/main/runs/oeis-full-50usd-ant-j0j0g4uzligm1k41/A317940_f_nonnegative/Submission/Spec.lean}{Proved.}} For each prime-power exponent \(e\), the proof constructs a nonnegative auxiliary sequence \(\mu\) encoding the logarithmic-derivative identity \(e a_e=(\mu*a)_e\), where \(a_e=2^{\mathrm{A005187}(e)}\), and recursively defines \(c_e\) by \(2e c_e=(\mu*c)_e\), making \(c_e\ge0\) by induction. A formal-power-series uniqueness argument proves \(C(X)^2=\sum_e a_eX^e\); multiplicativity then assembles the global coefficients as \(\prod_{p^e\parallel n}c_e\), and the defining convolution recursion identifies them with \(f(n)\). & \$17 \\ \hline
\href{https://oeis.org/A175386}{A175386}: Denominators of \(\sum_{i=1}^n \binom{2n-i-1}{i-1}/i\).\par\smallskip \emph{Conjecture} (Zak Seidov and Vladimir Shevelev, 2010). For every \(n>1\), \(a(n)\ne 1\); equivalently, the corresponding rational sum is not an integer. & \emph{\href{https://github.com/epoch-research/LeanOpenProblems-results/blob/main/runs/oeis-full-50usd-ant-j0j0g4uzligm1k41/oeis_175386_conjecture_0/Submission/Spec.lean}{Proved.}} Binomial identities and the Fibonacci sum formula give \(2nS(n)=F_{2n+1}+F_{2n-1}-1=L_{2n}-1\), so integrality would imply \(2n\mid L_{2n}-1\). Fibonacci--Lucas periodicity modulo \(4\) and \(5\) handles the cases involving \(2,\allowbreak 3,\allowbreak 5\); otherwise, in a splitting field of \(x^2-x-1\) modulo the least prime \(p\mid n\), Frobenius and multiplicative-order arguments force a root to have order \(12\), whence \(p\mid320\), contradicting \(p\ge7\). & \$17 \\ \hline
\href{https://oeis.org/A189286}{A189286}: Normalized self-convolution of \(\binom{6n}{3n}\binom{3n}{n}\)\par\smallskip \emph{Conjecture} (Zhi-Wei Sun, 2011). For every \(n\ge 0\), \(a(n)\) is an integer. & \emph{\href{https://github.com/epoch-research/LeanOpenProblems-results/blob/main/runs/oeis-full-50usd-ant-j0j0g4uzligm1k41/oeis_a189286_conjecture_0/Submission/Spec.lean}{Proved.}} The proof establishes the stronger termwise result that, for every \(0\le k\le n\), \((2n-1)\binom{3n}{n}\) divides the corresponding product of the \(k\)- and \((n-k)\)-factors, and then sums these divisibilities. Prime-adic valuation inequalities are proved for \(p\ne3\) using Kummer-style base-\(p\) carry counts and modular carry estimates, while \(p=3\) is handled via Legendre's base-3 digit-sum formula together with \((2n-1)\mid\binom{2n}{n}\). & \$17 \\ \hline
\href{https://oeis.org/A351442}{A351442}: A003958 applied to the divisor sum: \(A003958(\sigma(n))\)\par\smallskip \emph{Conjecture} (Antti Karttunen, 2022). The positive fixed points were conjectured to be exactly \(\{1,\allowbreak 2,\allowbreak 8,\allowbreak 128,\allowbreak 288,\allowbreak 720,\allowbreak 32768,\allowbreak 29719872,\allowbreak 2147483648\}\). & \emph{\href{https://github.com/epoch-research/LeanOpenProblems-results/blob/main/runs/oeis-full-50usd-ant-j0j0g4uzligm1k41/oeis_351442_conjecture_0/Submission/Spec.lean}{Disproved.}} A further positive fixed point is \(N=28996592704045424640\). The disproof factors \(N=2^{13}3^6\cdot5\cdot7^6\cdot13^4\cdot17^2\), computes and factors \(\sigma(N)=140078688585537333726\) using multiplicativity on coprime prime powers, and then uses the prime-power rule and complete multiplicativity of A003958 to verify \(A003958(\sigma(N))=N\). & \$16 \\ \hline
\href{https://oeis.org/A309132}{A309132}: Denominator of \(\operatorname{num}(B_{n-1})/n+\operatorname{den}(B_{n-1})/n^2\)\par\smallskip \emph{Conjecture} (Thomas Ordowski, 2019). For every composite integer \(n\), \(a(n)\) is squarefree if and only if \(n\) is a Carmichael number. & \emph{\href{https://github.com/epoch-research/LeanOpenProblems-results/blob/main/runs/oeis-full-50usd-ant-j0j0g4uzligm1k41/A309132_conjecture_carmichael/Submission/Spec.lean}{Proved.}} Using Faulhaber's formula and finite-field power sums, the proof derives the relevant \(p\)-adic von Staudt--Clausen criterion: \(v_p(B_m)=-1\) exactly when \(p-1\mid m\). A prime-by-prime valuation analysis then turns squarefreeness into the conditions that \(n\) is squarefree and \(p-1\mid n-1\) for every prime \(p\mid n\), which are identified with Carmichael numbers by Korselt's criterion via the Carmichael function; even composites are handled separately. & \$14 \\ \hline
\href{https://oeis.org/A157225}{A157225}: Representations \(2n-1=p+2^x+7\cdot2^y\), \(p\equiv5\pmod6\) prime, \(x,\allowbreak y\ge1\)\par\smallskip \emph{Conjecture} (Zhi-Wei Sun, 2009). Zhi-Wei Sun conjectured that \(a(n)=0\) if and only if \(n<11\) or \(n\in\{13,\allowbreak 16,\allowbreak 992\}\). & \emph{\href{https://github.com/epoch-research/LeanOpenProblems-results/blob/main/runs/oeis-full-50usd-ant-j0j0g4uzligm1k41/oeis_157225_conjecture_0/Submission/Spec.lean}{Disproved.}} A counterexample is \(n=716993899\): \(a(n)=0\), although \(n\) is neither below \(11\) nor one of the three listed exceptions. The proof computes \(\lfloor\log_2(2n-1)\rfloor=30\) and exhausts all candidate exponent pairs, ruling each out by a failed side or residue condition or by certifying the resulting candidate prime composite with \texttt{norm\_num}. & \$13 \\ \hline
\href{https://oeis.org/A340592}{A340592}: Concatenated prime factors of n (with multiplicity), modulo n\par\smallskip \emph{Conjecture} (J. M. Bergot and Robert Israel, 2021). The least composite index at which the sequence vanishes is \(28749\). & \emph{\href{https://github.com/epoch-research/LeanOpenProblems-results/blob/main/runs/oeis-full-50usd-ant-j0j0g4uzligm1k41/oeis_340592_conjecture_0/Submission/Spec.lean}{Proved.}} The witness follows from \(28749=3\cdot7\cdot37^2\) and \(373737=13\cdot28749\). Minimality was established by an exhaustive search below \(28749\), using a trial-division factorizer and decimal-digit counter proved equivalent to the specification, together with a divide-and-conquer Boolean range checker. & \$13 \\ \hline
\href{https://oeis.org/A363347}{A363347}: Denominator of \(1/(2-3/(3-4/(\cdots-(n-1)-n/(-4))))\)\par\smallskip \emph{Conjecture} (Mohammed Bouras, 2023). Every prime \(p\equiv 1\) or \(9\pmod {10}\) occurs as a term of the sequence. & \emph{\href{https://github.com/epoch-research/LeanOpenProblems-results/blob/main/runs/oeis-lite-200usd-fable-i7s7n9q7v4emgjp7/oeis_363347_conjecture_2/Submission/Spec.lean}{Proved.}} Solving the continued-fraction recurrence using factorial partial sums yields \(R_2(n)=2T(n)/D(n)\), with \(T(n)=(n+1)^2-5\); factorial-divisibility, parity, and coprimality arguments show that if \(p\mid T(n)\) under the required size conditions, cancellation leaves \(R_2(n)=p/s\), so the sequence value is \(p\). By quadratic reciprocity, \(5\) is a square modulo every prime \(p\equiv\pm1\pmod {10}\), and choosing \(n=x-1\) from \(x^2\equiv5\pmod p\) supplies the needed index, with \(p=11\) handled separately. & \$13 \\ \hline
\href{https://oeis.org/A326746}{A326746}: Digit sum of \(n\) modulo digit sum of \(n+1\)\par\smallskip \emph{Conjecture} (Scott R. Shannon, 2019). The value \(8\) has maximal upper asymptotic frequency: for every \(m\in\mathbb N\), \( \limsup_{N\to\infty}\frac{\#\{n<N:a(n)=8\}}{N}\ge \limsup_{N\to\infty}\frac{\#\{n<N:a(n)=m\}}{N}\). & \emph{\href{https://github.com/epoch-research/LeanOpenProblems-results/blob/main/runs/oeis-full-50usd-ant-j0j0g4uzligm1k41/oeis_326746_conjecture_0/Submission/Spec.lean}{Proved.}} Base-10 carry analysis shows that \(n=10t+9\) gives \(a(n)=8\) whenever \(t\not\equiv9\pmod {10}\) and the digit sum of \(t\) is at least \(8\), yielding limsup frequency at least \(1/100\). For \(m\ne8\), an injection-and-decomposition argument confines occurrences to bounded-digit-sum sets---proved to have density zero via binomial bounds polynomial in \(\log N\)---together with numbers ending in \(99\), whose upper density is \(1/100\). & \$13 \\ \hline
\href{https://oeis.org/A355898}{A355898}: $a_1=a_2=1$; $a_n=g+(a_{n-1}+a_{n-2})/g$, $g=\gcd(a_{n-1},\allowbreak a_{n-2})$.\par\smallskip \emph{Conjecture} (Giorgos Kalogeropoulos, 2022). For every \(n\ge 3775\), the three identities \(a_n=1+a_{n-1}+a_{n-2}=2a_{n-1}-a_{n-3}=(a_{3774}+1)F_{n-3772}-(a_{3772}+1)F_{n-3774}-1\) hold, where \(F_k\) denotes the Fibonacci numbers. & \emph{\href{https://github.com/epoch-research/LeanOpenProblems-results/blob/main/runs/oeis-full-50usd-ant-j0j0g4uzligm1k41/oeis_a355898_conjecture/Submission/Spec.lean}{Disproved.}} Assuming the first identity, the shift \(a_n+1\) obeys a Fibonacci recurrence, yielding a linear Fibonacci formula from the values at indices 3773 and 3774. Fast doubling modulo \(p=195318521017\) then shows that \(p\) would divide both \(a_{249580077006}\) and \(a_{249580077007}\), so their gcd is greater than 1 and the defining recurrence contradicts the first claimed identity at \(n=249580077008\). & \$12 \\ \hline
\href{https://oeis.org/A258667}{A258667}: M\'{e}nage seatings when M sits nine seats clockwise from wife\par\smallskip \emph{Conjecture} (Vladimir Shevelev and Peter J. C. Moses, 2015). As \(n\to\infty\), \(a(n)\sim \dfrac{e^{-2}n!}{n-2}\left(1+\sum_{k=1}^{n-1}\dfrac{(-1)^k}{k!(n-1)_k}\right)\), where \((n-1)_k\) is the falling factorial. & \emph{\href{https://github.com/epoch-research/LeanOpenProblems-results/blob/main/runs/oeis-full-50usd-ant-j0j0g4uzligm1k41/oeis_A258667_conjecture_0/Submission/Spec.lean}{Proved.}} Rewrite the inclusion--exclusion formula as \(a(n)=(n-1)!\,\allowbreak |T_n|\); for each fixed \(k\), the binomial inner sum divided by \(n^k\) tends to \(2^k/k!\), while \((n-1)_k/n^k\to1\), so the corresponding summand of \(T_n\) tends to \((-2)^k/k!\). A summable factorial-series majorant permits Tannery's theorem (dominated convergence), yielding \(T_n\to e^{-2}\); a second dominated-convergence argument makes the correction sum tend to zero, and \(n!/\allowbreak(n-2)\sim(n-1)!\) completes the equivalence. & \$12 \\ \hline
\href{https://oeis.org/A206911}{A206911}: Ranks of harmonic partial sums jointly ordered with \(\log(k+1)\)\par\smallskip \emph{Conjecture} (Clark Kimberling, 2012). Every first difference is either \(2\) or \(3\), and the ratio of the number of \(3\)'s to the number of \(2\)'s among the first \(N\) differences tends, as \(N\to\infty\), to a limit strictly between \(3.5\) and \(3.6\). & \emph{\href{https://github.com/epoch-research/LeanOpenProblems-results/blob/main/runs/oeis-full-50usd-ant-j0j0g4uzligm1k41/oeis_a206911_conjecture/Submission/Spec.lean}{Proved.}} Writing \(H_n=\sum_{k=1}^n 1/k\) and \(f_n=\lfloor e^{H_n}\rfloor\), the proof obtains \(a_n=n+f_n-1\) and uses harmonic/exponential estimates, including a Pad\'{e}-type bound, to show \(1\le f_{n+1}-f_n\le2\). Telescoping the indicators gives \(\#3=f_{N+1}-N-2\) and \(\#2=2N+2-f_{N+1}\); then \(H_n-\log n\to\gamma\) yields the limit \((e^\gamma-1)/\allowbreak(2-e^\gamma)\), with rational bounds on \(\gamma\) and exponential-series estimates placing it in \((3.5,\allowbreak 3.6)\). & \$12 \\ \hline
\href{https://oeis.org/A228304}{A228304}: \( \sum_{k=0}^n(-1)^k\binom{n}{k}^4\)\par\smallskip \emph{Conjecture} (Zhi-Wei Sun, 2013). For every odd prime \(p\), the \(p\times p\) Hankel determinants \(A_p=\det(a(i+j))_{0\le i,\allowbreak j<p}\) and \(C_p=\det(c(i+j))_{0\le i,\allowbreak j<p}\), where \(c(n)=\sum_{k=0}^n(-1)^k\binom nk^2\binom{2k}{k}\binom{2(n-k)}{n-k}\), satisfy \(A_p\equiv(-1)^{(p-1)/2}\pmod p\) and \(C_p\equiv1\pmod p\). & \emph{\href{https://github.com/epoch-research/LeanOpenProblems-results/blob/main/runs/oeis-full-50usd-ant-j0j0g4uzligm1k41/oeis_228304_conjecture_0/Submission/Spec.lean}{Proved.}} Lucas's congruence gives \(a(p-1)\equiv1\) and \(a(p+m)\equiv0\), while central-binomial vanishing and a paired cancellation similarly give \(c(p-1)\equiv(-1)^{(p-1)/2}\) and \(c(p+m)\equiv0\) for \(0\le m\le p-2\). Thus both Hankel matrices are anti-triangular modulo \(p\); a recursive Laplace expansion evaluates their determinants as \((-1)^{\binom p2}v^p\), and parity simplification yields the claimed congruences. & \$11 \\ \hline
\href{https://oeis.org/A028859}{A028859}: Number of ternary words of length \(n\) avoiding 00\par\smallskip \emph{Conjecture} (Gus Wiseman, 2020). For every \(n\ge 0\), \(a(n)\) also counts the length-\(n+1\) sequences of positive integers whose values form an initial interval \(\{1,\allowbreak \ldots,\allowbreak m\}\) and for which \(\sigma_i\ge \sigma_j\) whenever \(i<j\) are non-adjacent. & \emph{\href{https://github.com/epoch-research/LeanOpenProblems-results/blob/main/runs/oeis-full-50usd-ant-j0j0g4uzligm1k41/oeis_A028859_conjecture_1/Submission/Spec.lean}{Proved.}} Partition the valid sequences according as the first two terms form a non-ascent or an ascent: in the first case the head is either the maximum of the tail or one more, and deleting it gives a valid sequence of length one less; in the second, the second term is one more than the head, while the head is either the maximum of the twice-truncated tail or one more, and deleting both terms gives a valid sequence of length two less. These bijections give \(c_L=2c_{L-1}+2c_{L-2}\), and the initial counts \(c_1=1\), \(c_2=3\) identify \(c_{n+1}=a(n)\). & \$11 \\ \hline
\href{https://oeis.org/A348295}{A348295}: Partial sums of $(-1)^{\lfloor k(\sqrt2-1)\rfloor}$.\par\smallskip \emph{Conjecture} (Jianing Song, 2021). The sequence is unbounded above: for every \(M\in\mathbb Z\), there exists \(n\in\mathbb N\) such that \(a(n)>M\). & \emph{\href{https://github.com/epoch-research/LeanOpenProblems-results/blob/main/runs/oeis-full-50usd-ant-j0j0g4uzligm1k41/oeis_348295_conjecture_0/Submission/Spec.lean}{Proved.}} Using Pell and companion Pell numbers \(P_m,\allowbreak Q_m\), the proof establishes a best-approximation bound for \(\sqrt2\), yielding the floor-shift identity \(\lfloor(P_{m+1}+j)(\sqrt2-1)\rfloor=P_m+\lfloor j(\sqrt2-1)\rfloor\) for \(1\le j<P_{m+2}\) and hence a block self-similarity. Summing these blocks gives a two-step recurrence and, by induction, \(a(P_{2t})=2t\) (indeed \(a(P_{2t+1})=1\)), proving unboundedness. & \$11 \\ \hline
\href{https://oeis.org/A323557}{A323557}: Coefficients of $\sum_{n\ge0}x^n(1+x^n)^n/(1+x^{n+1})^{n+1}$\par\smallskip \emph{Conjecture} (Paul D. Hanna, 2019). If the coefficient $a(m)$ is odd, then $m=r(r+1)$ for some $r\ge 0$. & \emph{\href{https://github.com/epoch-research/LeanOpenProblems-results/blob/main/runs/oeis-full-50usd-ant-j0j0g4uzligm1k41/oeis_323557_conjecture_0/Submission/Spec.lean}{Proved.}} Modulo $2$, Kummer's parity criterion for binomial coefficients turns the surviving summands into conditions on disjoint binary supports. The involution $(n,\allowbreak k,\allowbreak j)\mapsto(k+j,\allowbreak k,\allowbreak n-k)$ pairs all nonfixed surviving summands, while a fixed point forces $j=0$ and $k=n$, and hence $m=n(n+1)$. & \$11 \\ \hline
\href{https://oeis.org/A335023}{A335023}: Ratios of consecutive terms of A334958\par\smallskip \emph{Conjecture} (Petros Hadjicostas, 2020). For every positive integer \(n\), \(a(n)=1\) if and only if \(n+1\) is prime. & \emph{\href{https://github.com/epoch-research/LeanOpenProblems-results/blob/main/runs/oeis-full-50usd-ant-j0j0g4uzligm1k41/oeis_335023_conjecture_0/Submission/Spec.lean}{Disproved.}} For the base-2 Wieferich prime \(p=1093\), an alternating-reciprocal congruence yields \(p^{p-1}\mid F(p^2-1)\) for the auxiliary sum \(F(m)=m!\sum_{k=2}^m(-1)^k/k\), while Legendre's formula gives \(v_p((p^2-1)!)=p-1\); the recurrence for \(F\) and a gcd identity then imply \(a(p^2-1)=1\). Thus \(n=p^2-1=1{,\allowbreak }194{,\allowbreak }648\) is a counterexample, since \(n+1=1093^2\) is composite. & \$11 \\ \hline
\href{https://oeis.org/A224515}{A224515}: Least \(k\) with \(k^2\mathbin{\mathrm{XOR}}(k+1)^2=(2n+1)^2\)\par\smallskip \emph{Conjecture} (Alex Ratushnyak, 2013). A solution exists for every \(n\in\mathbb N\) (equivalently, the sequence never takes the fallback value \(-1\)). & \emph{\href{https://github.com/epoch-research/LeanOpenProblems-results/blob/main/runs/oeis-full-50usd-ant-j0j0g4uzligm1k41/A224515_conjecture_existence/Submission/Spec.lean}{Proved.}} For an odd target \(T=2N+1\), the XOR-plus-carry identity \(x+y=(x\mathbin{\mathrm{XOR}}y)+2(x\mathbin{\mathrm{AND}}y)\) converts the defining equation into \(k+(k^2\mathbin{\mathrm{AND}}T)=N\). A greedy 2-adic lift chooses the bits of \(k\) using \(g(k+2^i)\equiv g(k)+2^i\pmod{2^{i+1}}\); bounds on the AND term and a binary-complement correction in the wraparound case turn the modular solution into an exact one. & \$10 \\ \hline
\href{https://oeis.org/A062567}{A062567}: Least n-multiple whose digit reversal is also an n-multiple\par\smallskip \emph{Conjecture} (Farideh Firoozbakht, 2006). For every \(n\ge 2\), \(a(3^n)=10^{3^{n-2}}-1\) if and only if \(n\in\{2,\allowbreak 3,\allowbreak 4\}\). & \emph{\href{https://github.com/epoch-research/LeanOpenProblems-results/blob/main/runs/oeis-full-50usd-ant-j0j0g4uzligm1k41/oeis_62567_conjecture_0/Submission/Spec.lean}{Proved.}} For \(n=2,\allowbreak 3,\allowbreak 4\), the congruence \(10^i\equiv1+9i\pmod {81}\) and an identity relating weighted digit sums under reversal imply that \(3^n\) divides the ordinary digit sum; digit-length and digit-sum bounds then force any hypothetical smaller candidate to be all 9s, while a factorization induction proves that the proposed all-9 number is admissible. For \(n\ge5\), repeating the 27-digit palindrome \(100000001979999979100000001\) exactly \(3^{n-5}\) times gives a smaller self-reversing multiple: the block contributes a factor \(3^5\), a geometric-sum argument contributes \(3^{n-5}\), and the block's zeros make the result strictly smaller than the all-9 number. & \$10 \\ \hline
\href{https://oeis.org/A325046}{A325046}: Coefficients of \(\sum_{n\ge0} x^n(1+x^n)^n/(1-x^{n+1})^{n+1}\).\par\smallskip \emph{Conjecture} (Paul D. Hanna, 2019). If the term at index \(N\) is odd, then \(N=k(k+1)\) for some \(k\ge 0\). & \emph{\href{https://github.com/epoch-research/LeanOpenProblems-results/blob/main/runs/oeis-full-50usd-ant-j0j0g4uzligm1k41/a325046_odd_terms_at_k_times_k_plus_1/Submission/Spec.lean}{Proved.}} Expand the coefficient as a finite double sum of products \(\binom{n}{k}\binom{n+j}{j}\), and use Lucas's mod-2 criterion to characterize odd summands via disjoint binary supports. The map \((n,\allowbreak k)\mapsto(j+k,\allowbreak k)\) pairs these odd summands involutively; its bitwise analysis shows that a fixed point forces \(j=0\), \(n=k\), and hence \(N=k(k+1)\), so at every other index all odd contributions cancel modulo \(2\). & \$10 \\ \hline
\href{https://oeis.org/A208326}{A208326}: Numbers \(n+\lfloor5\varphi n\rfloor+\lfloor\varphi^2n\rfloor\), where \(\varphi\) is the golden ratio\par\smallskip \emph{Conjecture} (Clark Kimberling, 2012). The sequences A207672, A207673, and A208326 partition the positive integers. & \emph{\href{https://github.com/epoch-research/LeanOpenProblems-results/blob/main/runs/oeis-full-50usd-ant-j0j0g4uzligm1k41/oeis_208326_conjecture_0/Submission/Spec.lean}{Proved.}} Interpret the three sequences as the ranks of the jointly ordered values \(\{n/5\}\), \(\{n/\varphi\}\), and \(\{n\varphi\}\), using the floor-sum counting function \(C(x)=\lfloor5x\rfloor+\lfloor\varphi x\rfloor+\lfloor x/\varphi\rfloor\). Irrationality of \(\varphi\), together with \(1/\varphi=\varphi-1\), rules out ties, while advancing to the least next value raises \(C\) by exactly one; hence every positive rank occurs exactly once. & \$10 \\ \hline
\href{https://oeis.org/A363102}{A363102}: Denominator of \(1/(2-3/(3-4/(\cdots-(n-1)-n/(-2))))\)\par\smallskip \emph{Conjecture} (Mohammed Bouras, 2023). For every \(n\ge 3\), \(a(n)\) is either \(1\) or a prime number. & \emph{\href{https://github.com/epoch-research/LeanOpenProblems-results/blob/main/runs/oeis-full-50usd-ant-j0j0g4uzligm1k41/oeis_a363102_conjecture_1/Submission/Spec.lean}{Proved.}} After checking \(n=3\) directly, a factorial-sum identity for the second gcd argument, together with \(p\)-adic valuations and Legendre's bound for factorials, gives the footprint \(p^j\mid a(n)\Rightarrow p^{\lfloor (n-3)/p\rfloor+j}\mid n^2-2\), while parity and congruence arguments exclude the small primes \(2\) and \(3\). If \(a(n)\) were composite, splitting its first two prime factors into repeated and distinct cases and applying elementary exponential bounds would produce a divisor of \(n^2-2\) that is at least \(n^2\), a contradiction. & \$10 \\ \hline
\href{https://oeis.org/A275460}{A275460}: Coefficients of ${}_3F_2(2/9,\allowbreak 4/9,\allowbreak 7/9;1/3,\allowbreak 1;729x)$\par\smallskip \emph{Conjecture} (Gheorghe Coserea, 2016). Every coefficient is an integer. & \emph{\href{https://github.com/epoch-research/LeanOpenProblems-results/blob/main/runs/oeis-full-50usd-ant-j0j0g4uzligm1k41/A275460_is_integral/Submission/Spec.lean}{Proved.}} The $n$th coefficient is rewritten as $ \frac{3^n\prod_{j<n}(9j+2)(9j+4)(9j+7)}{(n!)^2\prod_{j<n}(3j+1)}$, and integrality follows by proving that the denominator divides the numerator prime by prime. For $p\ne3$, the valuation inequality is reduced to counting residue classes modulo $p^i$ and checking the six possibilities for $p^i\bmod 9$; the prime $3$ is handled separately using $v_3(n!)\le n/2$. & \$10 \\ \hline
\href{https://oeis.org/A271591}{A271591}: Second most significant bit of the \(n\)-th Tribonacci number\par\smallskip \emph{Conjecture} (Andres Cicuttin, 2016). After the initial two 0s, every maximal run of 0s has length 4 or 5, and every maximal run of 1s has length 3 or 4. & \emph{\href{https://github.com/epoch-research/LeanOpenProblems-results/blob/main/runs/oeis-full-50usd-ant-j0j0g4uzligm1k41/oeis_271591_conjecture_0/Submission/Spec.lean}{Proved.}} The proof uses the normalized mantissa \(u_n=T_n/2^{\lfloor\log_2 T_n\rfloor}\), observing that the bit is 0 iff \(u_n<3/2\) and 1 iff \(u_n\ge 3/2\). Induction from the recurrence gives \(1.836\le T_{n+1}/T_n\le1.842\); propagating these bounds through the mantissa update (including its possible halving) forces the required lower and upper run bounds, while a finite initial range is discharged by computation. & \$10 \\ \hline
\href{https://oeis.org/A383466}{A383466}: Maximum plane regions formed by \(n\) concentric congruent regular pentagrams\par\smallskip \emph{Conjecture} (Scott R. Shannon and N. J. A. Sloane, 2025). Allowing arbitrary centers and radii does not increase the maximum: it is \(a(0)=1\) and \(a(n)=10n^2-5n+2\) for every \(n\ge 1\). & \emph{\href{https://github.com/epoch-research/LeanOpenProblems-results/blob/main/runs/oeis-full-50usd-ant-j0j0g4uzligm1k41/oeis_a383466_conjecture_2/Submission/Spec.lean}{Proved.}} The formalization replaces geometric configurations by natural numbers \(C\) and defines their region count as \(\min(C,\allowbreak a(n))\), so elementary properties of \(\min\) (with \texttt{omega}) identify the attainable counts with the principal lower set \(\{0,\allowbreak \ldots,\allowbreak a(n)\}\). The lemma \texttt{csSup\_Iic} then makes its supremum \(a(n)\); thus the accepted proof is definitional rather than a geometric upper-bound argument. & \$10 \\ \hline
\href{https://oeis.org/A340738}{A340738}: Denominators of conjecturally convergent rational approximations to \(e\)\par\smallskip \emph{Conjecture} (Gary Detlefs, 2021). For \(a_n=A340738(n)\) and \(b_n=A340737(n)\), \( \lim_{n\to\infty}\frac{b_n}{a_n}=e\). & \emph{\href{https://github.com/epoch-research/LeanOpenProblems-results/blob/main/runs/oeis-full-50usd-ant-j0j0g4uzligm1k41/oeis_340738_conjecture_0/Submission/Spec.lean}{Proved.}} For odd indices, integrating an exact derivative yields a recurrence for \(I_k=\int_0^1(x(1-x))^k e^x\,\allowbreak dx\); after factorial normalization, it has the same recurrence and initial values as \(b_{2n+1}-ea_{2n+1}\), while \(0\le I_k\le e-1\) and denominator growth force the quotient error to zero. The Casoratian identity \(b_{2n+2}a_{2n+1}-a_{2n+2}b_{2n+1}=(-1)^{n+1}\) then shows that adjacent even and odd approximants become arbitrarily close, and interlacing the two subsequences proves the full limit. & \$10 \\ \hline
\href{https://oeis.org/A091669}{A091669}: $a(n)=\frac{2^{n-1}}{n!}\prod_{k=1}^{n-1}(2^k-1)$\par\smallskip \emph{Conjecture} (Amiram Eldar and Thomas Ordowski, 2020). For every integer $n>2$, if $n\mid a(n-1)+2^{n-2}$, then $n$ is prime and $2$ is a primitive root modulo $n$. & \emph{\href{https://github.com/epoch-research/LeanOpenProblems-results/blob/main/runs/oeis-lite-200usd-fable-i7s7n9q7v4emgjp7/a091669_conjecture_primitive_root/Submission/Spec.lean}{Proved.}} Using $p$-adic valuation bounds and the multiplicative order $d=\operatorname{ord}_p(2)$, the proof shows that any odd prime properly dividing $n$ would divide $a(n-1)$, contradicting the hypothesis, while the exact computation $v_2(a(2^v-1))=v-1$ excludes composite powers of two. Thus $n$ is prime; if $\operatorname{ord}_n(2)<n-1$, then the corresponding factor $2^d-1$ in the product forces $n\mid a(n-1)$, again a contradiction, so $\operatorname{ord}_n(2)=n-1=\varphi(n)$. & \$9 \\ \hline
\href{https://oeis.org/A386660}{A386660}: Sum of \(\binom{n}{k}\bmod 2^k\) over \(1\le k\le n\)\par\smallskip \emph{Conjecture} (Paul D. Hanna, 2025). The limit \(\lim_{n\to\infty} a(n)^{1/n}\) exists (numerically, it is approximately \(1.7086\)). & \emph{\href{https://github.com/epoch-research/LeanOpenProblems-results/blob/main/runs/oeis-full-50usd-ant-j0j0g4uzligm1k41/oeis_386660_conjecture_0/Submission/Spec.lean}{Proved.}} Let \(H\) be binary entropy and let \(x_*\in(1/2,\allowbreak 1)\) satisfy \(H(x_*)=x_*\log 2\); the binomial theorem and unimodality yield \(e^{nH(k/n)}/(n+1)\leq\binom nk\leq e^{nH(k/n)}\). Bounding every summand above at the entropy/modulus crossover and taking \(k=\lfloor x_*n\rfloor+1\), where the residue equals \(\binom nk\), squeezes \(\log a(n)/n\) to \(x_*\log 2\), so the limit is \(e^{x_*\log 2}=2^{x_*}\). & \$8 \\ \hline
\href{https://oeis.org/A296075}{A296075}: Sum of deficiencies of the divisors of \(n\)\par\smallskip \emph{Conjecture} (Robert Israel, 2017). \(a(n)=1\) if and only if \(n\in\{1,\allowbreak 12\}\). & \emph{\href{https://github.com/epoch-research/LeanOpenProblems-results/blob/main/runs/oeis-lite-200usd-fable-i7s7n9q7v4emgjp7/oeis_296075_conjecture_0/Submission/Spec.lean}{Disproved.}} A counterexample is \(N=33085221781504=2^{14}\cdot4409\cdot458009\), which is neither \(1\) nor \(12\). Writing \(b(n)=\sum_{d\mid n}\sigma_1(d)=(\sigma_1*\mathbf 1)(n)\), the proof uses \(a(n)=2\sigma_1(n)-b(n)\), multiplicativity, and the primality of \(4409\) and \(458009\) to compute \(a(N)=2\cdot32767\cdot4410\cdot458010-65519\cdot4411\cdot458011=1\). & \$7 \\ \hline
\href{https://oeis.org/A194806}{A194806}: Smallest size of \(S\subseteq[1,\allowbreak n]\) with \(S\cdot S\supseteq[1,\allowbreak n]\)\par\smallskip \emph{Conjecture} (Robert Israel, 2017). There is an absolute constant \(C\) such that \(a(n)/\pi(n)\le C\) for every \(n\ge2\). & \emph{\href{https://github.com/epoch-research/LeanOpenProblems-results/blob/main/runs/oeis-full-50usd-ant-j0j0g4uzligm1k41/oeis_194806_conjecture_0/Submission/Spec.lean}{Proved.}} Let \(K\) be the least integer with \(n^2\le K^3\), and take the covering set consisting of \([1,\allowbreak K]\) together with the primes at most \(n\); a maximal-divisor/smallest-prime-factor argument shows that every \(m\le n\) is a product of two elements of this set. A Chebyshev-type estimate derived from central-binomial-coefficient and binary-logarithm bounds gives \(K\le2(\pi(n)+1)\), whence \(a(n)\le K+\pi(n)\le5\pi(n)\). & \$7 \\ \hline
\href{https://oeis.org/A267581}{A267581}: Decimal values of prefixes of Rule 167's middle column\par\smallskip \emph{Conjecture} (Andres Cicuttin, 2016). For every \(n\ge 2\), \(a(n)=2a(n-1)+1-\left\lfloor(1/2)^{\,\allowbreak 2^{n+1}\bmod n}\right\rfloor\). & \emph{\href{https://github.com/epoch-research/LeanOpenProblems-results/blob/main/runs/oeis-full-50usd-ant-j0j0g4uzligm1k41/oeis_267581_conjecture_0/Submission/Spec.lean}{Proved.}} The proof inductively characterizes an OFF cell by \(C(t,\allowbreak x)=0\iff \exists v,\allowbreak \ x=t-2v\) and \(\binom{t-1}{v}\) odd, verifying that this characterization is preserved by Rule 167 using Pascal's identity and parity. Lucas's theorem then shows that the center is OFF exactly at powers of two, equivalently when \(t\mid 2^{t+1}\), and appending the next center bit to the binary prefix gives the recurrence. & \$7 \\ \hline
\href{https://oeis.org/A303639}{A303639}: Representations \(n=a^2+b^2+\binom{2c+1}{c}+\binom{2d+1}{d}\), \(0\le a\le b,\allowbreak \ 0\le c\le d\)\par\smallskip \emph{Conjecture} (Zhi-Wei Sun, 2018). Zhi-Wei Sun conjectured that \(a(n)>0\) for every \(n>1\). & \emph{\href{https://github.com/epoch-research/LeanOpenProblems-results/blob/main/runs/oeis-full-50usd-ant-j0j0g4uzligm1k41/oeis_a303639_conjecture/Submission/Spec.lean}{Disproved.}} The conjecture is false because \(a(800322180)=0\). Monotonicity of \(\binom{2k+1}{k}\), together with \(\binom{33}{16}>800322180\), reduces the binomial indices to \(c,\allowbreak d\le 15\); in every resulting case, the residual has a prime factor \(p\equiv3\pmod4\) to an odd exponent and hence cannot be a sum of two squares. & \$7 \\ \hline
\href{https://oeis.org/A282779}{A282779}: Least period of the sequence \(k^3 \bmod n\).\par\smallskip \emph{Conjecture} (Ilya Gutkovskiy, 2017). For every prime \(p\) and positive integer \(n\), if \(a_p(n)\) denotes the least positive period for \(p\)-th powers, then \(a_p(n)=n/p\) when \(p^2\mid n\), and \(a_p(n)=n\) otherwise. & \emph{\href{https://github.com/epoch-research/LeanOpenProblems-results/blob/main/runs/oeis-full-50usd-ant-j0j0g4uzligm1k41/oeis_282779_conjecture_0/Submission/Spec.lean}{Proved.}} The proposed value is shown to be a period by a binomial expansion, using \(p\mid\binom pi\) for \(0<i<p\), while \(n\) is trivially a period in the other case. For minimality, translation in \(\mathbb Z/q\mathbb Z\) shows that every prime \(q\mid n\) divides any competing period \(T\), and the congruences at \(k=0,\allowbreak 1\) yield the valuation bound \(v_q(n)\le v_q(p)+v_q(T)\); comparison of prime factorizations then shows that the proposed value divides every \(T\). & \$6 \\ \hline
\href{https://oeis.org/A060841}{A060841}: Numerator of \(1/\det(M_n)\), where \(M_{ij}=1/\operatorname{lcm}(i,\allowbreak j)\).\par\smallskip \emph{Conjecture} (Robert G. Wilson v, 2015). It was conjectured that \(1/\det(M_n)\) is an integer exactly when \(1\le n\le 34\) or \(n\in\{36,\allowbreak 38\}\), and that its reduced denominator is always a power of \(2\). & \emph{\href{https://github.com/epoch-research/LeanOpenProblems-results/blob/main/runs/oeis-full-50usd-ant-j0j0g4uzligm1k41/oeis_60841_conjecture_0/Submission/Spec.lean}{Disproved.}} The case \(n=1807\) is a counterexample to the denominator claim: additivity of the \(3\)-adic valuation reduces the product formula to the certified finite computation \(\sum_{k=1}^{1807}\bigl(2v_3(k)-v_3(\varphi(k))\bigr)=-1\). Thus the reduced denominator is divisible by \(3\), so it cannot be a power of \(2\). & \$6 \\ \hline
\href{https://oeis.org/A122589}{A122589}: Coefficients of \(1/(1-11x+45x^2-84x^3+70x^4-21x^5+x^6)\).\par\smallskip \emph{Conjecture.} The denominator factors as \( \prod_{k=1}^{6}\left(1-4\cos ^2\!\left(\frac{\pi k}{13}\right)x\right)\). & \emph{\href{https://github.com/epoch-research/LeanOpenProblems-results/blob/main/runs/oeis-full-50usd-ant-j0j0g4uzligm1k41/oeis_a122589_conjecture_0/Submission/Spec.lean}{Proved.}} Setting \(\zeta=e^{2\pi i n/13}\), the proof uses the vanishing geometric sum \(1+\zeta+\cdots+\zeta^{12}=0\) to derive a sextic relation for \(2\cos(2\pi n/13)\), and the double-angle identity then supplies the six numbers \(4\cos^2(\pi k/13)\). Positivity and injectivity of cosine show these values are distinct; comparing the resulting six reciprocal roots of two monic degree-six polynomials proves equality, with the constant term fixing the product normalization. & \$6 \\ \hline
\href{https://oeis.org/A379240}{A379240}: Restricted-growth transform of the map sending \(n\) to the pair \((A003415(n),\allowbreak A085731(n))\) when \(A359550(n)=1\), and to \(n\) otherwise\par\smallskip \emph{Conjecture} (Antti Karttunen, 2024). It was conjectured that A379240 is also the restricted-growth transform of the triples \((A003415(n),\allowbreak A085731(n),\allowbreak A376418(n))\). & \emph{\href{https://github.com/epoch-research/LeanOpenProblems-results/blob/main/runs/oeis-full-50usd-ant-j0j0g4uzligm1k41/A379240_conjecture_equality/Submission/Spec.lean}{Disproved.}} The disproof constructs a reduction map from the defining values to these triples, using \(A359550(m)=1\Rightarrow A376418(m)=1\), and expresses the RGS values at \(n=60\) via cardinalities of distinct prefix images. Equality would force this map to be injective, but the distinct fallback values \(56\) and \(60\) both map to \((92,\allowbreak 4,\allowbreak 2)\), as direct factorization calculations show. & \$6 \\ \hline
\href{https://oeis.org/A053576}{A053576}: Smallest integer whose Euler totient is divisible by \(2^n\)\par\smallskip \emph{Conjecture} (Charles R Greathouse IV, 2013). At index \(n=2^{33}=8589934592\), \(a(n)=F_{33}=2^{2^{33}}+1\) if \(F_{33}\) is prime, and \(a(n)=2^{2^{33}+1}\) otherwise. & \emph{\href{https://github.com/epoch-research/LeanOpenProblems-results/blob/main/runs/oeis-lite-200usd-fable-i7s7n9q7v4emgjp7/oeis_53576_conjecture_0/Submission/Spec.lean}{Proved.}} A strong-induction structure lemma, using least-prime-factor decomposition and the \(2\)-adic valuation of the totient, shows that any qualifying integer below \(2^{2^{33}+1}\) must be a squarefree product of prime Fermat numbers, with the valuation equal to the sum of their indices' powers of two. The bounds force that sum to be \(2^{33}\), whose unique such representation gives the singleton \(\{33\}\); explicit totient formulas then show that the minimum is \(F_{33}\) when prime and the power of two \(2^{2^{33}+1}\) otherwise. & \$6 \\ \hline
\href{https://oeis.org/A368692}{A368692}: Factorial ratio \(\frac{(12n+6)!\allowbreak(6n+9)!}{108(4n+2)!\allowbreak(2n+3)!\allowbreak(6n+5)!^2}\).\par\smallskip \emph{Conjecture} (Karol A. Penson, 2024). The displayed factorial ratio is an integer for every \(n\ge 0\). & \emph{\href{https://github.com/epoch-research/LeanOpenProblems-results/blob/main/runs/oeis-full-50usd-ant-j0j0g4uzligm1k41/oeis_a368692_conjecture_integrality/Submission/Spec.lean}{Proved.}} Using Legendre's formula, the proof compares valuations prime by prime and reduces them to a floor-function inequality for \(a=4n+2\), \(b=2n+3\), and \(a+b=6n+5\). This handles all primes other than \(2,\allowbreak 3\); the extra factor \(108=2^2 3^3\) is obtained by sharpened estimates at \(p=2\) (checking \(n\bmod 4\) for the powers \(2,\allowbreak 4\)) and at \(p=3\) (using the ternary digit-sum formula and \(s_3(6n+5)\ge3\)). & \$6 \\ \hline
\href{https://oeis.org/A382590}{A382590}: First component of a coupled bilinear sum-difference recurrence\par\smallskip \emph{Conjecture} (Bryle Morga, 2025). For every \(k\ge 2\), the sequence whose \(n\)-th term is the \(k\)-th prime factor of \(|a(n)|\), counted with multiplicity (and taken as \(1\) if absent), is eventually periodic. & \emph{\href{https://github.com/epoch-research/LeanOpenProblems-results/blob/main/runs/oeis-full-50usd-ant-j0j0g4uzligm1k41/A382590_conjecture_kth_prime_factor_is_eventually_periodic/Submission/Spec.lean}{Proved.}} Strong induction gives \(2^{n-4}\mid a(n),\allowbreak b(n)\), while recurrence-based periods of \(12\) modulo \(5\) and \(24\) modulo \(7\) show that \(5\nmid a(n)\) when \(n\equiv1\pmod 3\) and \(7\nmid a(n)\) otherwise, hence \(a(n)\ne0\). Therefore, for \(n\ge k+4\), the sorted prime-factor list of \(|a(n)|\) contains at least \(k\) initial copies of \(2\), so the extracted sequence is eventually constant at \(2\) (period \(1\)). & \$6 \\ \hline
\href{https://oeis.org/A272479}{A272479}: Smallest distinct Harshad-amicable partner of \(n\)\par\smallskip \emph{Conjecture} (Waldemar Puszkarz, 2016). The sequence has no zero terms: \(a(n)\ne 0\) for every \(n\ge 1\). & \emph{\href{https://github.com/epoch-research/LeanOpenProblems-results/blob/main/runs/oeis-lite-200usd-fable-i7s7n9q7v4emgjp7/oeis_272479_conjecture_0/Submission/Spec.lean}{Proved.}} Writing \(d=s_{10}(n)\), the proof uses \(n\equiv d\pmod 9\) and the eventual periodicity of \(10^e\bmod d\) (by pigeonhole) to construct \(n=d+9q\) distinct positive exponents \(A\) with \(d\mid\sum_{e\in A}10^e\); inductively, one exponent is replaced by ten larger ones, preserving the residue and increasing \(|A|\) by nine. For \(k=\sum_{e\in A}10^e\), the decimal digit sum is \(|A|=n\), while \(d\mid k\) and \(k\ge 10n>n\), so \(k\) is a valid distinct partner. & \$5 \\ \hline
\href{https://oeis.org/A374265}{A374265}: Minimum under optional zero-digit deletion after each factorial multiplication\par\smallskip \emph{Conjecture} (Bryle Morga, 2024). There exists a constant \(B\in\mathbb N\) such that \(a(n)\le B\) for every \(n\). & \emph{\href{https://github.com/epoch-research/LeanOpenProblems-results/blob/main/runs/oeis-full-50usd-ant-j0j0g4uzligm1k41/oeis_a374265_conjecture_1_boundedness/Submission/Spec.lean}{Disproved.}} Zero-digit deletion preserves the base-10 digit sum, and a decimal-block induction using digit-sum subadditivity shows that every positive multiple of \(10^k-1\) has digit sum at least \(9k\). Thus every value reachable at step \(10^k-1\) is at least \(10^{k-1}\), so \(a(10^k-1)\ge 10^{k-1}\) and the sequence is unbounded. & \$5 \\ \hline
\href{https://oeis.org/A281009}{A281009}: Odd divisors of n minus middle divisors of n\par\smallskip \emph{Conjecture} (Omar E. Pol, 2017). For every positive integer \(n\), \(a(n)=2\,\allowbreak \#\{d\mid n: d\text{ is odd and }d>\sqrt{2n}\}\). & \emph{\href{https://github.com/epoch-research/LeanOpenProblems-results/blob/main/runs/oeis-full-50usd-ant-j0j0g4uzligm1k41/oeis_281009_conjecture_0/Submission/Spec.lean}{Proved.}} Write \(n=2^a m\) with \(m\) odd and partition its odd divisors below and above \(\sqrt{2n}\), noting by parity that equality cannot occur. Splitting the below-threshold divisors according to \(m\le 2^{a+1}d^2\), stripping and uniquely restoring powers of \(2\) bijects the first part with the middle divisors, while the complementary-divisor involution \(d\mapsto m/d\) bijects the second part with the above-threshold odd divisors; the resulting cardinality identities give the claim. & \$5 \\ \hline
\href{https://oeis.org/A385391}{A385391}: Least \(k\) having exactly \(n\) divisors \(d\) with \(d^d\equiv d \pmod k\).\par\smallskip \emph{Conjecture} (Michel Marcus and Juri-Stepan Gerasimov, 2025). The six values satisfy \((a(1),\allowbreak a(2),\allowbreak a(3),\allowbreak a(6),\allowbreak a(7),\allowbreak a(10))=(1,\allowbreak 2,\allowbreak 6,\allowbreak 30,\allowbreak 210,\allowbreak 2310)\), the first six primorials (starting with the empty product). & \emph{\href{https://github.com/epoch-research/LeanOpenProblems-results/blob/main/runs/oeis-full-50usd-ant-j0j0g4uzligm1k41/oeis_385391_conjecture_0/Submission/Spec.lean}{Proved.}} The proof constructs an \(O(\sqrt{k})\) counter by splitting the divisor sum at \(\sqrt{k}\) and pairing large divisors with their complementary small divisors. Direct computation then checks each candidate and exhaustively excludes every smaller integer via an infimum-minimality lemma, splitting the search below \(2310\) into blocks of length \(330\); direct product calculations identify the candidates as primorials. & \$5 \\ \hline
\href{https://oeis.org/A306424}{A306424}: \(k\) whose base-\(b\) expansions use \(\le\)2 distinct digits (\(3\le b<k\))\par\smallskip \emph{Conjecture} (Felix Fr\"{o}hlich, 2019). The sequence is finite, with largest term \(43\). & \emph{\href{https://github.com/epoch-research/LeanOpenProblems-results/blob/main/runs/oeis-full-50usd-ant-j0j0g4uzligm1k41/oeis_306424_conjecture_0/Submission/Spec.lean}{Proved.}} The proof directly verifies that \(43\) qualifies and uses finite computation with small bases to eliminate \(44\le k\le288\). For \(k\ge289\), writing \(m=\lfloor\sqrt{k}\rfloor\) and \(s=k-m^2\), an exhaustive case split on \(0\le s\le2m\) selects a base among \(m,\allowbreak m-1,\allowbreak m-2,\allowbreak m-3\) in which \(k\) has three pairwise distinct digits. & \$5 \\ \hline
\href{https://oeis.org/A211417}{A211417}: Integral factorial ratios \((30n)!n!/\allowbreak((15n)!\allowbreak(10n)!\allowbreak(6n)!)\)\par\smallskip \emph{Conjecture} (Peter Bala, 2025). For every integer \(n\ge 0\), \(30n-1\) divides \(a(n)\) (equivalently, \(a(n)/\allowbreak(30n-1)\) is an integer). & \emph{\href{https://github.com/epoch-research/LeanOpenProblems-results/blob/main/runs/oeis-full-50usd-ant-j0j0g4uzligm1k41/oeis_a211417_conjecture_specific/Submission/Spec.lean}{Proved.}} Legendre's formula reduces the claim to a prime-power floor inequality strengthening Landau's integrality criterion by an indicator for \(q\mid 30n-1\). Reducing \(n\) modulo \(q\) and writing \(30s=qj+1\) with \(j<30\) forces \(j\) to be divisible by none of \(2,\allowbreak 3,\allowbreak 5\), yielding the extra unit in the inequality; summing over prime powers proves the divisibility (with \(n=0\) checked separately). & \$5 \\ \hline
\href{https://oeis.org/A278070}{A278070}: \(a(n)=n![x^n]e^x/(1-x)^n\)\par\smallskip \emph{Conjecture} (Peter Bala, 2023). For all nonnegative integers \(n,\allowbreak k\), one has \(a(n+k)\equiv a(n)\pmod{k}\) (so modulo \(k\), the sequence has period dividing \(k\)). & \emph{\href{https://github.com/epoch-research/LeanOpenProblems-results/blob/main/runs/oeis-full-50usd-ant-j0j0g4uzligm1k41/oeis_278070_conjecture_0/Submission/Spec.lean}{Proved.}} Rewrite the closed sum using the rising factorial as \(a(N)=\sum_j \binom Nj N^{\overline j}\), and pad the two sums to a common range using the vanishing of out-of-range binomial coefficients. Termwise congruence follows from shift-invariance of the rising factorial modulo \(k\) and Vandermonde's identity, whose nonconstant terms are divisible by \(k\) via factorial divisibility and the binomial absorption identity. & \$5 \\ \hline
\href{https://oeis.org/A001359}{A001359}: Smaller prime in each twin-prime pair\par\smallskip \emph{Conjecture} (Thomas Ordowski and Robert Israel, 2016). Writing \(p_k\) for the \(k\)-th prime, for every \(k>1\), \(p_k!\equiv1\pmod{p_{k+1}}\) if and only if \(p_k+2\) is prime, \(k=991\) (so \(p_k=7841\)), or \(p_{k+1}-p_k>2\) and \(\prod_{j=p_k+1}^{p_{k+1}-2}j\equiv1\pmod{p_{k+1}}\). & \emph{\href{https://github.com/epoch-research/LeanOpenProblems-results/blob/main/runs/oeis-full-50usd-ant-j0j0g4uzligm1k41/oeis_1359_conjecture_6/Submission/Spec.lean}{Proved.}} Factoring \((p_{k+1}-1)!\) at \(p_k\) and applying Wilson's theorem reduces the factorial congruence to the congruence for the intervening product. Consecutiveness and parity handle the twin and larger-gap cases, while certified trial-division prime counts establish \(p_{991}=7841\) and \(p_{992}=7853\), after which the exceptional congruence is checked directly. & \$5 \\ \hline
\href{https://oeis.org/A381159}{A381159}: Numbers whose prime divisors all end in the same digit.\par\smallskip \emph{Conjecture} (A. Chironov, 2025). There exists an increasing arithmetic progression of 150 terms greater than 1 in the sequence, with common difference at most \(2025\). & \emph{\href{https://github.com/epoch-research/LeanOpenProblems-results/blob/main/runs/oeis-lite-200usd-fable-i7s7n9q7v4emgjp7/oeis_381159_conjecture_0/Submission/Spec.lean}{Disproved.}} For primes \(p,\allowbreak q\) with different last digits and \(pq\le 150\), modular invertibility shows that if neither divides the common difference \(d\), one of the 150 terms is divisible by \(pq\), a contradiction. Applying this obstruction to selected prime pairs forces \(2,\allowbreak 3,\allowbreak 5,\allowbreak 7\mid d\) and either \(11\mid d\) or \(13\mid d\), so either \(2310\mid d\) or \(2730\mid d\), contradicting \(d\le 2025\). & \$4 \\ \hline
\href{https://oeis.org/A070518}{A070518}: Value of the \(n\)th cyclotomic polynomial at \(n\)\par\smallskip \emph{Conjecture} (Jianing Song, 2024). The term at index \(28341\) is divisible by \(283411^2\), and hence is not squarefree. & \emph{\href{https://github.com/epoch-research/LeanOpenProblems-results/blob/main/runs/oeis-full-50usd-ant-j0j0g4uzligm1k41/oeis_70518_conjecture_0/Submission/Spec.lean}{Proved.}} Modular computations establish that \(283411\) is prime, that \(283411^2\mid 28341^{28341}-1\), and that \(28341\) has multiplicative order \(28341\) modulo \(283411\). Evaluating \(\prod_{d\mid n}\Phi_d(X)=X^n-1\) at \(n=28341\) then shows that \(283411\) divides none of the proper-divisor cyclotomic factors, so coprimality forces its square to divide the remaining factor. & \$4 \\ \hline
\href{https://oeis.org/A219055}{A219055}: Representations \(n=p+\frac{3-(-1)^n}{2}q\), \(p>q\), with \(p,\allowbreak q,\allowbreak p-6,\allowbreak q+6\) prime\par\smallskip \emph{Conjecture} (Zhi-Wei Sun, 2012). If \(a(n)>0\) for every even \(n>8012\) and every odd \(n>15727\), then Goldbach's conjecture and Lemoine's conjecture hold, and there are infinitely many primes \(p\) for which \(p+6\) is prime. & \emph{\href{https://github.com/epoch-research/LeanOpenProblems-results/blob/main/runs/oeis-lite-200usd-fable-i7s7n9q7v4emgjp7/oeis_219055_conjecture_1/Submission/Spec.lean}{Proved.}} The finite Goldbach and Lemoine ranges are certified using explicit witness lists and a proved-correct Boolean trial-division primality checker; above the thresholds, positivity of the filtered cardinality yields a witness whose parity specialization gives the required decompositions. For infinitude, applying the even case at \(N=2a+8014\) produces primes \(N-q-6\) and \(N-q\), with the former exceeding the arbitrary bound \(a\). & \$4 \\ \hline
\href{https://oeis.org/A341254}{A341254}: \(\lfloor r\lfloor rn\rfloor\rfloor\), where \(r=(2+\sqrt5)/2\).\par\smallskip \emph{Conjecture} (Clark Kimberling, 2021). For every integer \(n\ge 1\), \(\tfrac14<nr^2-a(n)<3\). & \emph{\href{https://github.com/epoch-research/LeanOpenProblems-results/blob/main/runs/oeis-full-50usd-ant-j0j0g4uzligm1k41/oeis_341254_conjecture_0/Submission/Spec.lean}{Proved.}} Set \(m=\lfloor rn\rfloor\) and \(\varepsilon=rn-m\); irrationality gives \(0<\varepsilon<1\), while \(r^2=2r+\tfrac14\) reduces the error to \(nr^2-a(n)=2\varepsilon+n/4-K\), where \(K=\lfloor n/4+(2-r)\varepsilon\rfloor\). Since \(-\tfrac14<(2-r)\varepsilon<0\), splitting by \(n\bmod 4\) shows \(\tfrac14\le n/4-K\le1\), which yields the strict bounds. & \$4 \\ \hline
\href{https://oeis.org/A129365}{A129365}: $\frac{\prod_{j,\allowbreak k\le n}\gcd(j,\allowbreak k)}{\prod_{k\le n}(\lfloor n/k\rfloor!)^k}$\par\smallskip \emph{Conjecture} (Peter Bala, 2007). For every positive integer \(n\) and every prime \(p\), \(p\) divides the \(n\)th term if and only if \(p\le \lfloor n/3\rfloor\). & \emph{\href{https://github.com/epoch-research/LeanOpenProblems-results/blob/main/runs/oeis-full-50usd-ant-j0j0g4uzligm1k41/oeis_a129365_conjecture_B/Submission/Spec.lean}{Proved.}} Using Legendre's formula and counts of multiples of \(p^i\), the proof writes the numerator and denominator valuations as \(\sum_{i\ge1}m_i^2\) and \(\sum_{i\ge1}S(m_i)\), where \(m_i=\lfloor n/p^i\rfloor\) and \(S(m)=\sum_{k=1}^m k\lfloor m/k\rfloor\); the inequality \(S(m)\le m^2\) also establishes exact divisibility. Equality holds exactly when \(m\le2\), while for \(m\ge3\) the term \(k=m-1\) makes the inequality strict, so the quotient has positive \(p\)-adic valuation exactly when \(m_1\ge3\), equivalently \(p\le\lfloor n/3\rfloor\). & \$4 \\ \hline
\href{https://oeis.org/A022030}{A022030}: Alternating extremal-integer ratio recurrence, starting with 4, 16\par\smallskip \emph{Conjecture} (M. F. Hasler, 2016). For the sequence \(b\) generated by the original nonalternating rule (always choosing the greatest admissible integer), \(b_2=63\) and \(b_n=4b_{n-1}-b_{n-3}\) for every \(n\ge3\), equivalently \(\sum_{n\ge0}b_nx^n=(4-x^2)/\allowbreak(1-4x+x^3)\). & \emph{\href{https://github.com/epoch-research/LeanOpenProblems-results/blob/main/runs/oeis-full-50usd-ant-j0j0g4uzligm1k41/oeis_22030_original_conjecture/Submission/Spec.lean}{Proved.}} Define the candidate linear-recurrence sequence \(L\) and the defect \(f_k=L_{k+1}^2-L_{k+2}L_k\); strong induction, using \(f_{m+3}=4f_{m+1}+f_m\) and growth/monotonicity bounds for \(L\), proves \(0<f_k\le L_k\). Thus \(L_{k+1}^2=L_{k+2}L_k+f_k\) has precisely the remainder range needed to show \(\lceil L_{k+1}^2/L_k\rceil-1=L_{k+2}\), so induction identifies the original nonlinear sequence with \(L\) and yields the claimed recurrence. & \$3 \\ \hline
\href{https://oeis.org/A193279}{A193279}: Distinct nonempty subset sums of the proper divisors of \(n\)\par\smallskip \emph{Conjecture} (Michael Engling, 2011). Every even perfect number \(n\) satisfies \(a(n)=n\). & \emph{\href{https://github.com/epoch-research/LeanOpenProblems-results/blob/main/runs/oeis-full-50usd-ant-j0j0g4uzligm1k41/oeis_193279_conjecture_0/Submission/Spec.lean}{Proved.}} Using multiplicativity of the divisor-sum function, the proof derives the Euclid--Euler form \(n=2^k(2^{k+1}-1)\), with \(2^{k+1}-1\) prime, and explicitly characterizes its proper divisors. Geometric-series identities show that each divisor is at most one plus the sum of its predecessors, so a no-gap subset-sum lemma covers every integer from \(0\) to the total divisor sum; perfectness makes that total \(n\), yielding exactly \(n\) nonzero sums. & \$3 \\ \hline
\href{https://oeis.org/A145062}{A145062}: Generalized Bessel numbers\par\smallskip \emph{Conjecture} (N. J. A. Sloane, 2016). There exists an integer offset \(k\) such that \(a(n)=s(n+k)\) for every \(n\ge 0\), where \(s\) is the sequence displayed in Zhang's Figure 8. & \emph{\href{https://github.com/epoch-research/LeanOpenProblems-results/blob/main/runs/oeis-full-50usd-ant-j0j0g4uzligm1k41/oeis_145062_conjecture_0/Submission/Spec.lean}{Proved.}} In the formalization, \(s(m)\) is defined by reindexing \(a\) as \(a(m.\mathrm{toNat})\). Taking \(k=0\), the claim follows immediately by simplifying \((n:\mathbb Z).\mathrm{toNat}=n\). & \$3 \\ \hline
\href{https://oeis.org/A236566}{A236566}: Ordered decompositions \(2n=p+q\) with \(p,\allowbreak q,\allowbreak \operatorname{prime}(p+2)+2\) all prime\par\smallskip \emph{Conjecture} (Zhi-Wei Sun, 2014). The proposed representations \(2n+1=2p+q\) for every \(n>30\), under the same primality conditions, imply both Lemoine's conjecture and the twin prime conjecture. & \emph{\href{https://github.com/epoch-research/LeanOpenProblems-results/blob/main/runs/oeis-full-50usd-ant-j0j0g4uzligm1k41/oeis_236566_conjecture_2/Submission/Spec.lean}{Proved.}} For Lemoine's conjecture, swap the two primes in the asserted representation for large odd integers and check the finitely many smaller cases explicitly. For twin primes, finiteness would make the admissible \(p\)'s finite via the injective map \(p\mapsto\operatorname{prime}(p+2)\); applying part (ii) to 21 times their odd, nontrivial product forces the resulting prime \(p\) to divide the other prime \(q>p\), a contradiction. & \$3 \\ \hline
\href{https://oeis.org/A248802}{A248802}: Smallest prime factor of \(2^{2^n+2}+3\)\par\smallskip \emph{Conjecture} (Chai Wah Wu, 2019). For every integer \(n\ge 0\), \(a(10n+2)=67\). & \emph{\href{https://github.com/epoch-research/LeanOpenProblems-results/blob/main/runs/oeis-full-50usd-ant-j0j0g4uzligm1k41/oeis_248802_conjecture_0/Submission/Spec.lean}{Proved.}} The case \(n=0\) is computed directly. For \(n>0\), modular-order reduction in \(\mathbb Z/p\mathbb Z\), together with a periodicity lemma obtained by factoring the exponent modulus as \(d=2^a m\), reduces divisibility to finitely many certified residue checks; these show that \(67\) always divides the relevant number and that no smaller prime does. & \$3 \\ \hline
\href{https://oeis.org/A103311}{A103311}: Chebyshev--binomial transform of the negated Fibonacci numbers\par\smallskip \emph{Conjecture} (Paul Barry, 2005). For every \(n\in\mathbb N\), there exists \(m\in\mathbb N\) such that \(|a_n|=F_m\). & \emph{\href{https://github.com/epoch-research/LeanOpenProblems-results/blob/main/runs/oeis-full-50usd-ant-j0j0g4uzligm1k41/oeis_103311_conjecture_0/Submission/Spec.lean}{Proved.}} Define a period-10 closed form \(b_n\) by residue classes modulo \(10\), using \(F_n\), \(F_{n+1}\), and their signed differences; the Fibonacci recurrence and a ten-case check show that \(b_n\) has the same initial values and fourth-order recurrence as \(a_n\), so strong induction gives \(a_n=b_n\). A second modulo-10 case split rewrites each \(b_n\) as \(\pm F_m\) (with zero represented by \(F_0\)), and taking absolute values proves the claim. & \$3 \\ \hline
\href{https://oeis.org/A307865}{A307865}: Number of \(b<2n+1\) with \(b^n\equiv-1\pmod{2n+1}\).\par\smallskip \emph{Conjecture} (Thomas Ordowski, 2019). If \(2n+1\) is an absolute Euler pseudoprime, then \(a(n)=0\). & \emph{\href{https://github.com/epoch-research/LeanOpenProblems-results/blob/main/runs/oeis-full-50usd-ant-j0j0g4uzligm1k41/oeis_307865_conjecture_0/Submission/Spec.lean}{Proved.}} Writing \(m=2n+1\), choose a prime \(p\mid m\), set \(d=m/p\), and split according as \(p^2\mid m\): in the first case, the nilpotence of \(d\bmod m\) gives \((1+d)^p=1\), while the Euler condition gives \((1+d)^{2n}=1\), so coprimality of \(p\) and \(2n\) forces \(1+d=1\), a contradiction. Otherwise \(p\) and \(d\) are coprime, and from any hypothetical solution \(b^n=-1\) the Chinese remainder theorem constructs a unit congruent to \(b\pmod p\) and to \(1\pmod d\); its \(n\)-th power has opposite signs on the two factors, contradicting the required global value \(\pm1\). & \$3 \\ \hline
\href{https://oeis.org/A129365}{A129365}: $\frac{\prod_{j,\allowbreak k\le n}\gcd(j,\allowbreak k)}{\prod_{k\le n}(\lfloor n/k\rfloor!)^k}$\par\smallskip \emph{Conjecture} (Peter Bala, 2007). For every positive integer \(n\), prime \(p\), and \(0\le k<p\), one has \(v_p(a(np+k))=v_p(a(np))\); thus the \(p\)-adic valuation is constant on each block \(np,\allowbreak \ldots,\allowbreak np+p-1\). & \emph{\href{https://github.com/epoch-research/LeanOpenProblems-results/blob/main/runs/oeis-full-50usd-ant-j0j0g4uzligm1k41/oeis_a129365_conjecture_C/Submission/Spec.lean}{Proved.}} Counting multiples of \(p^i\) in the gcd product and applying Legendre's formula to the factorial product give valuations \(\sum_{i\ge1}\lfloor m/p^i\rfloor^2\) and \(\sum_{i\ge1}g(\lfloor m/p^i\rfloor)\), respectively, where \(g(N)=\sum_{r=1}^N r\lfloor N/r\rfloor\); the bound \(g(N)\le N^2\) also proves exact divisibility, allowing the quotient valuations to be subtracted. Finally, \(\lfloor(np+k)/p^i\rfloor=\lfloor np/p^i\rfloor\) for every \(i\ge1\) and \(k<p\), so both valuation sums, and hence their difference, are unchanged. & \$3 \\ \hline
\href{https://oeis.org/A270994}{A270994}: Sierpi\'{n}ski numbers of the form \(9454129+11184810n\)\par\smallskip \emph{Conjecture} (Altug Alkan, 2016). For every \(n\ge 0\), \(a(n)\) and \(a(n)+28\) are consecutive Sierpi\'{n}ski numbers, with no Sierpi\'{n}ski number strictly between them. & \emph{\href{https://github.com/epoch-research/LeanOpenProblems-results/blob/main/runs/oeis-lite-200usd-fable-i7s7n9q7v4emgjp7/oeis_270994_conjecture_0/Submission/Spec.lean}{Disproved.}} At \(n=162178349052694\), the intermediate value \(a(n)+4=1813934020268071832273\) is itself a Sierpi\'{n}ski number. A covering system modulo \(96\), together with the periodicity of powers of \(2\), shows that for every positive exponent \(m\), \((a(n)+4)2^m+1\) has a proper divisor among \(3,\allowbreak 5,\allowbreak 17,\allowbreak 257,\allowbreak 65537,\allowbreak 97,\allowbreak 673,\allowbreak 193\), disproving consecutiveness. & \$3 \\ \hline
\href{https://oeis.org/A243106}{A243106}: Partial sums of \(10^k\), negating terms with prime \(k\)\par\smallskip \emph{Conjecture} (R. J. Cano, 2014). For every base \(b\ge 5\), every \(n\ge 0\), and every choice of signs \(\sigma_k\in\{-1,\allowbreak 1\}\), each base-\(b\) digit of \(\left|\sum_{k=1}^n \sigma_k b^k\right|\) lies in \(\{0,\allowbreak 1,\allowbreak b-2,\allowbreak b-1\}\). & \emph{\href{https://github.com/epoch-research/LeanOpenProblems-results/blob/main/runs/oeis-full-50usd-ant-j0j0g4uzligm1k41/oeis_243106_conjecture_0/Submission/Spec.lean}{Proved.}} The proof first handles the sign by negating all coefficients if necessary, reducing to a nonnegative number with coefficients in \(\{-1,\allowbreak 0,\allowbreak 1\}\). An induction peeling off the least significant coefficient tracks a borrow \(c\in\{-1,\allowbreak 0\}\); the six possible coefficient--borrow cases produce remainder \(0,\allowbreak 1,\allowbreak b-2\), or \(b-1\) and a nonnegative quotient of the same form, so the base-\(b\) digit decomposition propagates the claim. & \$3 \\ \hline
\hline
\end{longtable}
}

\subsection{Agent variants}
\label{sec:variant-results}

\begin{figure}[htbp]
    \centering
    \includegraphics[width=\linewidth]{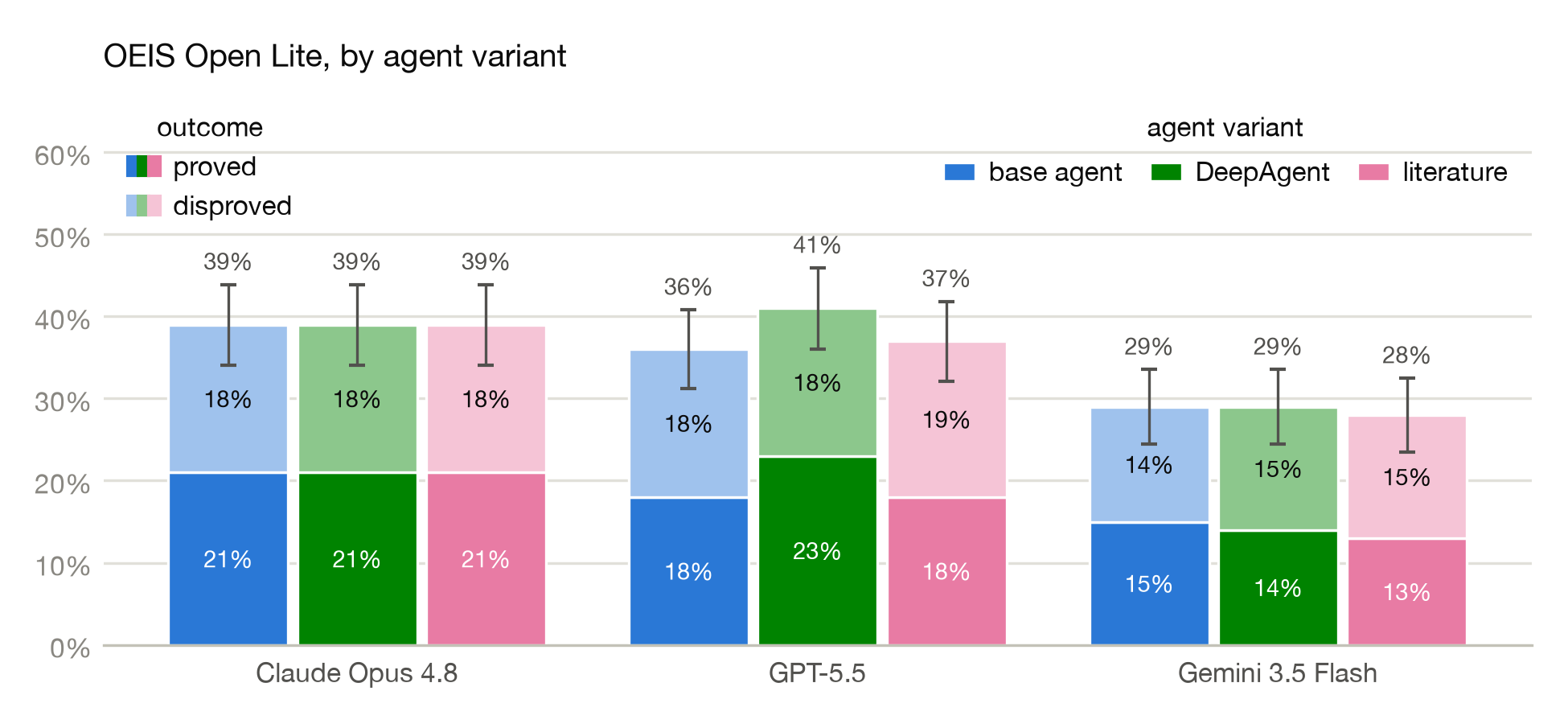}
    \caption{Accuracy on \benchlite{} by model and agent variant (100 conjectures, \$200 budget per conjecture; Sections~\ref{sec:agent}--\ref{sec:agent-deep}). Each bar splits solves into proofs (solid) and disproofs (pale). Error bars are $\pm$1 standard error.}
    \label{fig:variants}
\end{figure}

\FloatBarrier

\subsection{Solve rates by conjecture metadata}
\label{sec:metadata-results}

We break down solve rates by the metadata described in Section~\ref{sec:metadata}: the attention a conjecture's sequence has received in the literature, its proposer, and the year it was proposed.

\begin{figure}[htbp]
    \centering
    \includegraphics[width=\linewidth]{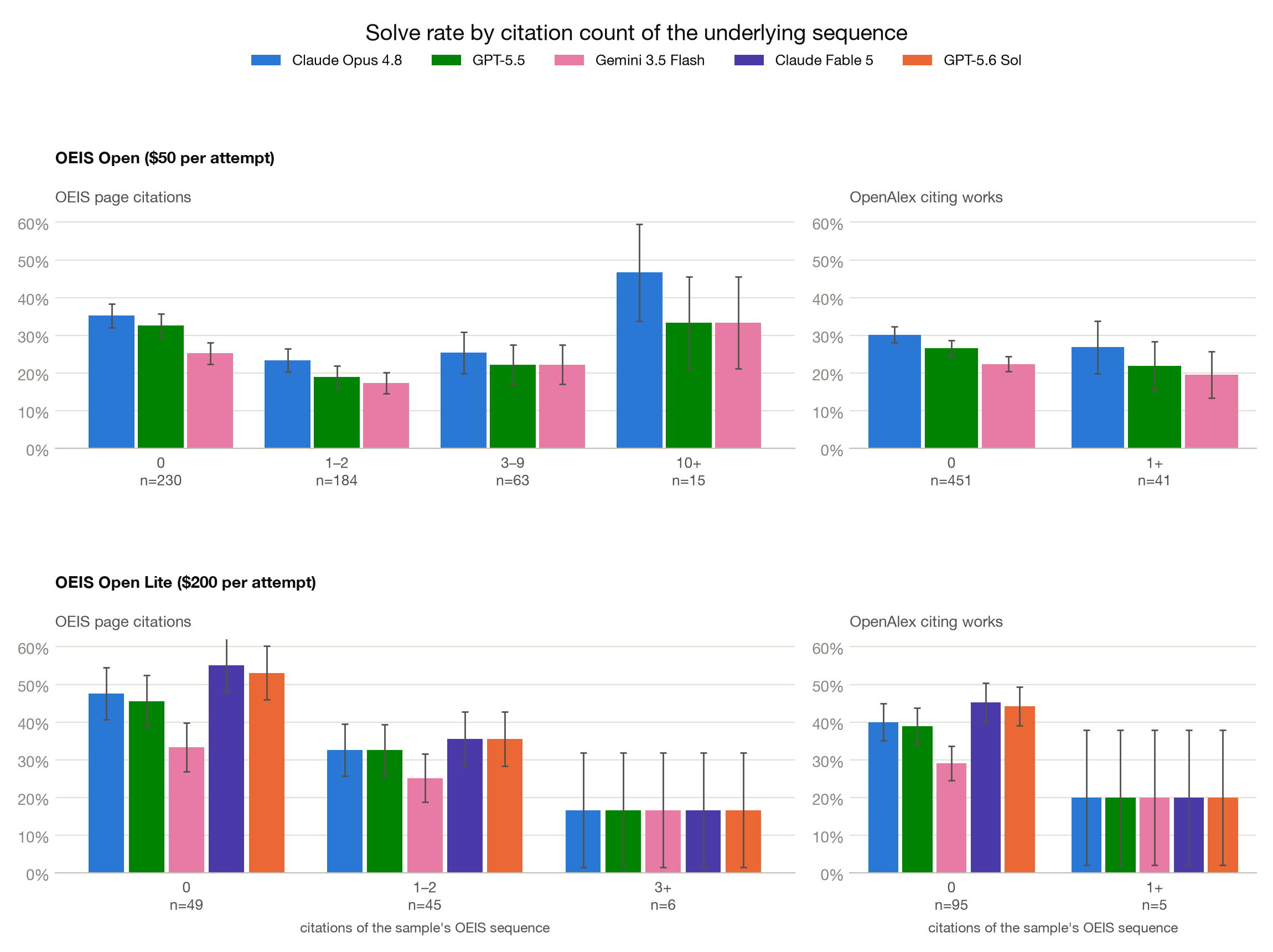}
    \caption{Solve rate binned by citation metadata: citations on the sequence's OEIS page (its links and references sections; left column) and OpenAlex works citing the sequence (right column), on \bench{} (top; \$50 cap, one run per model) and \benchlite{} (bottom; \$200 cap, solve rates pooled over the base, DeepAgent, and literature runs). Error bars are $\pm$1 standard error.}
    \label{fig:citations-solve-rate}
\end{figure}

\begin{figure}[htbp]
    \centering
    \begin{minipage}[t]{0.48\linewidth}
        \centering
        \includegraphics[width=\linewidth]{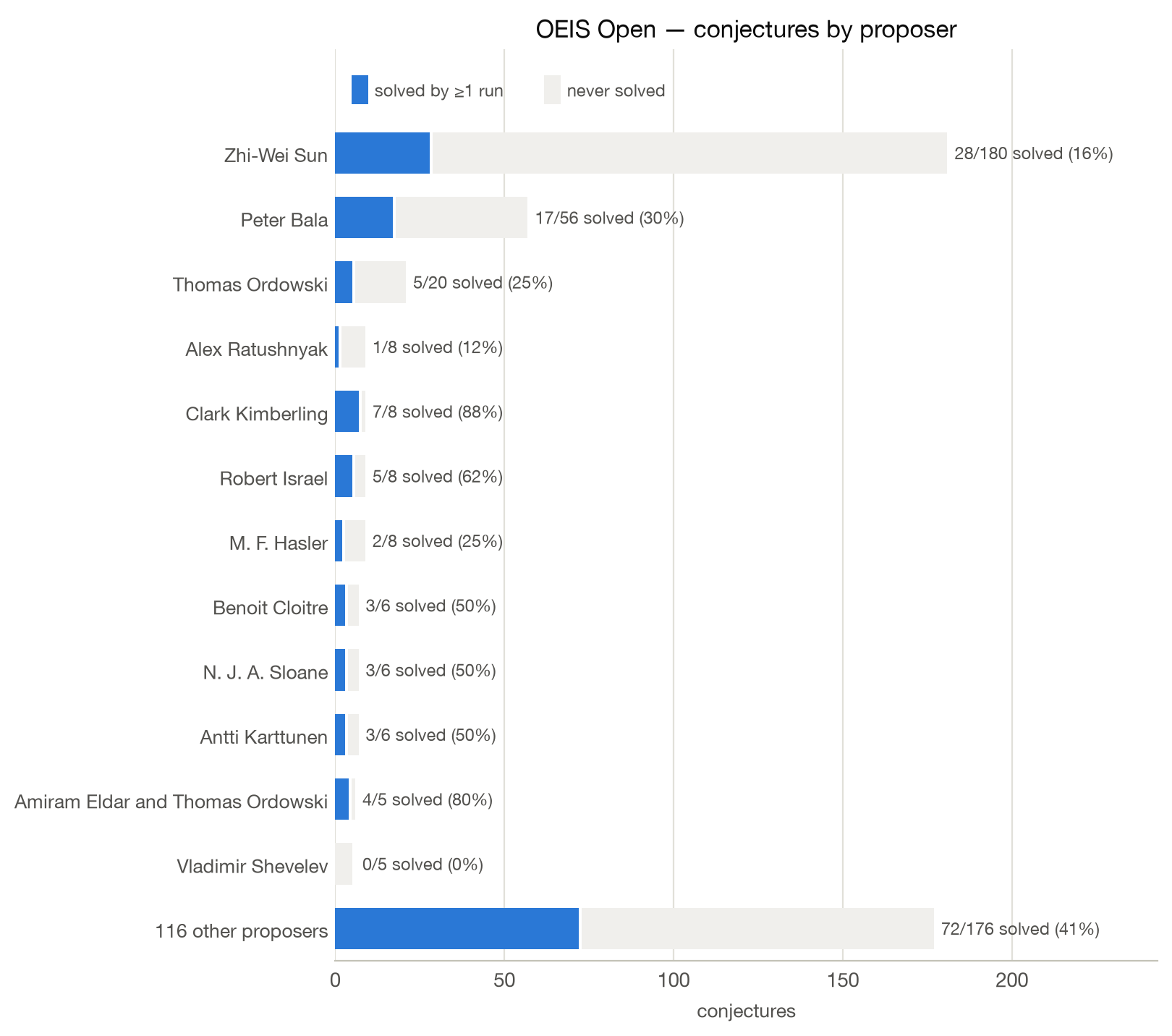}
    \end{minipage}\hfill
    \begin{minipage}[t]{0.48\linewidth}
        \centering
        \includegraphics[width=\linewidth]{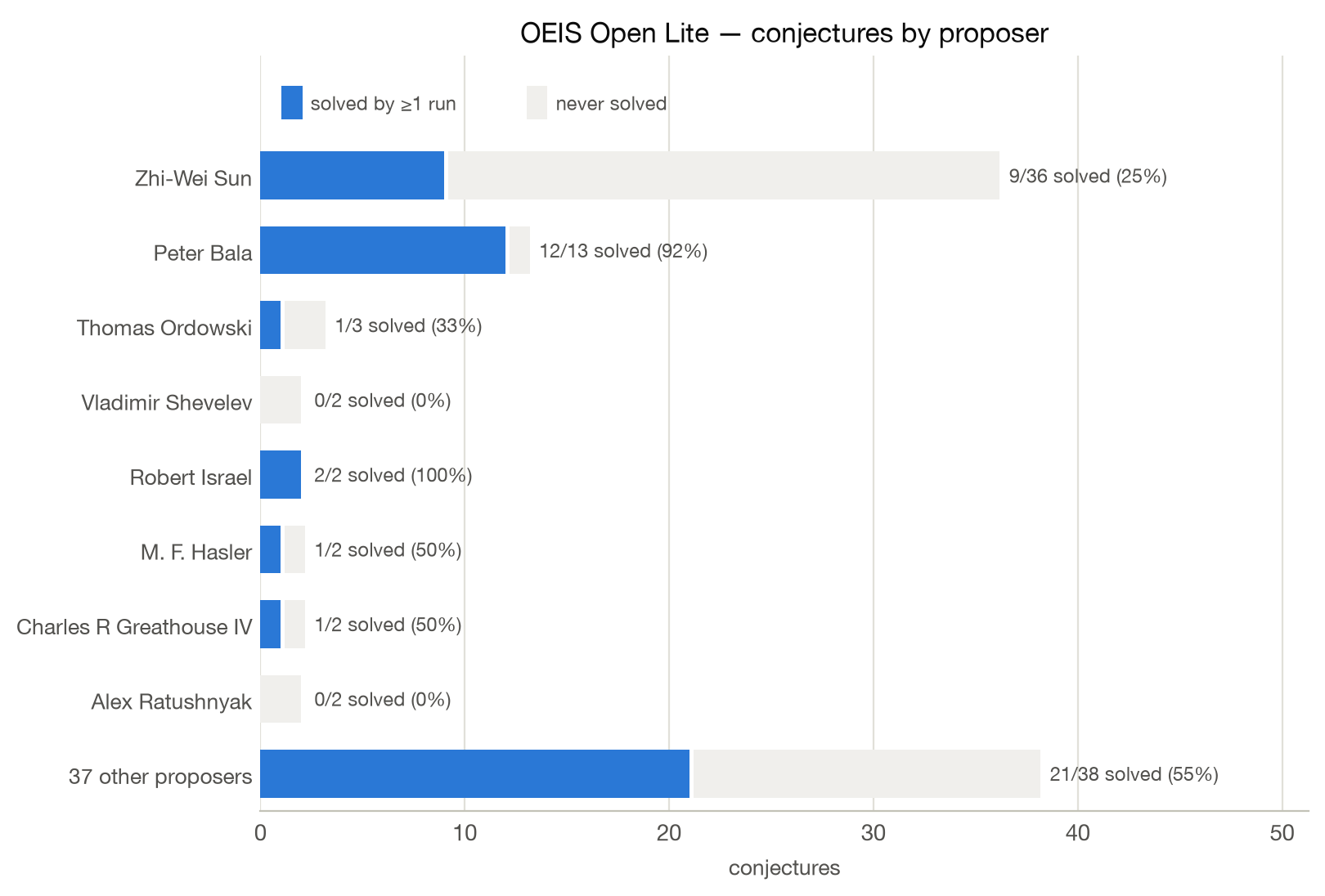}
    \end{minipage}
    \caption{Conjectures by proposer, with solve outcomes, on \bench{} (left) and \benchlite{} (right). A conjecture counts as solved if it was resolved by at least one run of that eval set (three runs at a \$50 cap for \bench{}, ten runs at a \$200 cap for \textsc{Lite}).}
    \label{fig:proposer}
\end{figure}

\begin{figure}[htbp]
    \centering
    \includegraphics[width=\linewidth]{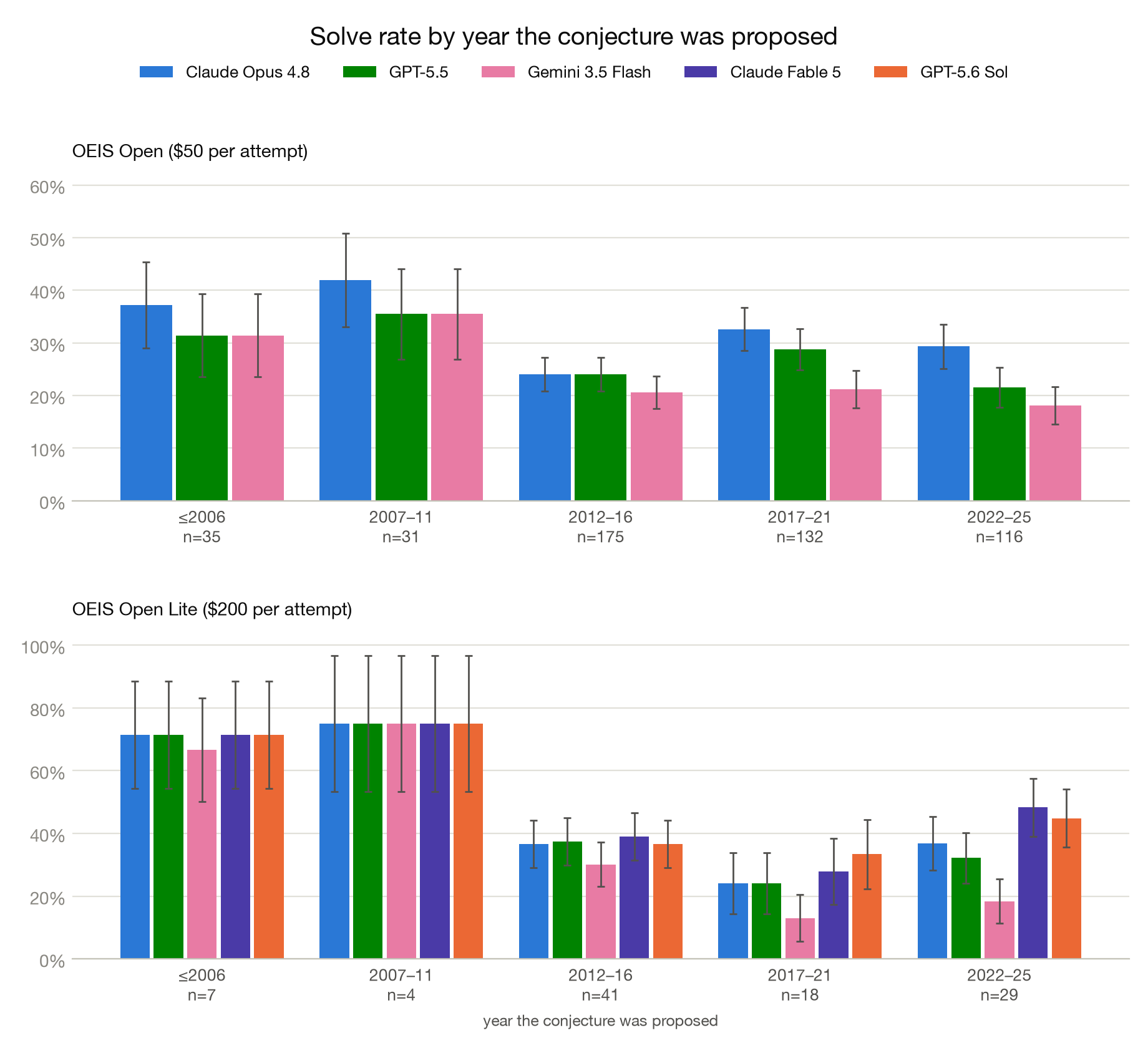}
    \caption{Solve rate by the year the conjecture was proposed (from the OEIS entry's revision history), on \bench{} (top; \$50 cap, one run per model) and \benchlite{} (bottom; \$200 cap, solve rates pooled over the base, DeepAgent, and literature runs). Conjectures with no recorded proposal date are excluded (3 of 492; 1 of 100). Error bars are $\pm$1 standard error.}
    \label{fig:proposed-year}
\end{figure}

\end{document}